\documentclass[letterpaper]{article} % DO NOT CHANGE THIS
\usepackage{aaai23}  % DO NOT CHANGE THIS
\usepackage{times}  % DO NOT CHANGE THIS
\usepackage{helvet}  % DO NOT CHANGE THIS
\usepackage{courier}  % DO NOT CHANGE THIS
\usepackage[hyphens]{url}  % DO NOT CHANGE THIS
\usepackage{graphicx} % DO NOT CHANGE THIS
\usepackage{natbib}  % DO NOT CHANGE THIS AND DO NOT ADD ANY OPTIONS TO IT
\usepackage{caption} % DO NOT CHANGE THIS AND DO NOT ADD ANY OPTIONS TO IT
\usepackage{algorithm}
\usepackage{algorithmic}

\usepackage{newfloat}
\usepackage{listings}
\DeclareCaptionStyle{ruled}{labelfont=normalfont,labelsep=colon,strut=off} % DO NOT CHANGE THIS
\floatstyle{ruled}
\newfloat{listing}{tb}{lst}{}
\floatname{listing}{Listing}
\nocopyright %-- Your paper will not be published if you use this command
\title{CowClip: Reducing CTR Prediction Model Training Time from 12 hours to 10 minutes on 1 GPU}
\author{
    Zangwei Zheng\textsuperscript{\rm 1}\thanks{Work done during an internship at Bytedance.} 
    \quad Pengtai Xu\textsuperscript{\rm 1}$^{*}$
    \quad Xuan Zou\textsuperscript{\rm 2}
    \quad Da Tang\textsuperscript{\rm 2}
    \quad Zhen Li\textsuperscript{\rm 2} \\
    Chenguang Xi\textsuperscript{\rm 2}
    \quad Peng Wu\textsuperscript{\rm 2}
    \quad Leqi Zou\textsuperscript{\rm 2}
    \quad Yijie Zhu\textsuperscript{\rm 2}
    \quad Ming Chen\textsuperscript{\rm 2} \\
    Xiangzhuo Ding\textsuperscript{\rm 2}
    \quad Fuzhao Xue\textsuperscript{\rm 1}
    \quad Ziheng Qin\textsuperscript{\rm 1}
    \quad Youlong Cheng\textsuperscript{\rm 2}
    \quad Yang You\textsuperscript{\rm 1}\thanks{Yang You is the corresponding author.}
}
\affiliations{
    \textsuperscript{\rm 1} Department of Computer Science, National University of Singapore\\
    \textsuperscript{\rm 2} Bytedance Inc.\\
    \{zangwei, pengtai, f-xue, zihengq, youy\}@comp.nus.edu.sg \\
    \{zouxuan, da.tang, zhen.li1, chenguang.xi, peng.wu, leqi.zou,\\
    yijie.zhu, ming.chen, xiangzhuo.ding, youlong.cheng\}@bytedance.com
}

\usepackage{url}            % simple URL typesetting
\usepackage{amsfonts}       % blackboard math symbols
\usepackage{nicefrac}       % compact symbols for 1/2, etc.
\usepackage{xcolor}         % colors

\usepackage{subcaption}
\usepackage{bm}
\usepackage{amsmath,amssymb}
\usepackage{booktabs}
\usepackage{multirow}
\usepackage[flushleft]{threeparttable}
\usepackage{makecell}

\newenvironment{packed_itemize}{
	\begin{itemize}
		\setlength{\itemsep}{0pt}
		\setlength{\parskip}{0pt}
		\setlength{\parsep}{0pt}
	}{\end{itemize}
}
\definecolor{f3green}{RGB}{0,144,81}
\definecolor{f3brown}{RGB}{148,82,0}
\newtheorem{scalingrule}{Scaling Rule}

\newcommand{\innerproduct}[2]{\langle #1, #2 \rangle}

\DeclareMathOperator*{\E}{\mathbb{E}}
\DeclareMathOperator*{\prob}{P}

\DeclareMathOperator*{\occur}{\texttt{count}}

\newcommand{\freq}{\prob(\text{id}_k^{\text{f}_j}\in \bm{x})}

\newcommand{\eg}{\textit{e}.\textit{g}.}

\usepackage[textsize=tiny]{todonotes}
\usepackage{hhline}
\usepackage{colortbl}
\definecolor{Gray}{gray}{0.9}
\definecolor{ForestGreen}{rgb}{0.13, 0.55, 0.13}
\definecolor{ForestGreen2}{rgb}{0.2, 0.5372549019607843, 0.1803921568627451}
\definecolor{newGREEN}{RGB}{34,139,34}
\definecolor{myMaroon}{RGB}{231, 52, 52}
\definecolor{Maroon}{rgb}{0.69, 0.19, 0.0}
\definecolor{my_cyan}{RGB}{112, 242, 244}

\usepackage{wrapfig}

\begin{document}

\maketitle

\begin{abstract}
The click-through rate (CTR) prediction task is to predict whether a user will click on the recommended item. As mind-boggling amounts of data are produced online daily, accelerating CTR prediction model training is critical to ensuring an up-to-date model and reducing the training cost. One approach to increase the training speed is to apply large batch training. However, as shown in computer vision and natural language processing tasks, training with a large batch easily suffers from the loss of accuracy. Our experiments show that previous scaling rules fail in the training of CTR prediction neural networks. To tackle this problem, we first theoretically show that different frequencies of ids make it challenging to scale hyperparameters when scaling the batch size. To stabilize the training process in a large batch size setting, we develop the adaptive Column-wise Clipping (CowClip). It enables an easy and effective scaling rule for the embeddings, which keeps the learning rate unchanged and scales the L2 loss. We conduct extensive experiments with four CTR prediction networks on two real-world datasets and successfully scaled 128 times the original batch size without accuracy loss. In particular, for CTR prediction model DeepFM training on the Criteo dataset, our optimization framework enlarges the batch size from 1K to 128K with over 0.1\% AUC improvement and reduces training time from 12 hours to 10 minutes on a single V100 GPU. Our code locates at \url{github.com/bytedance/LargeBatchCTR}.

% The click-through rate (CTR) prediction task is to predict whether a user will click on the recommended item. As mind-boggling amounts of data are produced online daily, accelerating CTR prediction model training is critical to ensuring an up-to-date model and reducing the training cost. One approach to increase the training speed is to apply large batch training. However, as shown in computer vision and natural language processing tasks, training with a large batch easily suffers from the loss of accuracy. Our experiments show that previous scaling rules fail in the training of CTR prediction neural networks. To tackle this problem, we first theoretically show that different frequencies of ids make it challenging to scale hyperparameters when scaling the batch size. To stabilize the training process in a large batch size setting, we develop the adaptive Column-wise Clipping (CowClip). It enables an easy and effective scaling rule for the embeddings, which keeps the learning rate unchanged and scales the L2 loss. We conduct extensive experiments with four CTR prediction networks on two real-world datasets and successfully scaled 128 times the original batch size without accuracy loss. In particular, for CTR prediction model DeepFM training on the Criteo dataset, our optimization framework enlarges the batch size from 1K to 128K with over 0.1% AUC improvement and reduces training time from 12 hours to 10 minutes on a single V100 GPU. Our code locates at github.com/bytedance/LargeBatchCTR.
\end{abstract}

\section{Introduction}

With the development of the Internet and the e-economy, numerous clicking happens in online shopping~\cite{Ma2020TemporalContextualRI, Zhou2019DeepIE}, video apps~\cite{GomezUribe2016TheNR,Xie2020KrakenMC} and web advertisements~\cite{Covington2016DeepNN,Zhao2019AIBoxCP}. Click-through Rate (CTR) prediction is to predict whether a user will click on the recommended item. It is a fundamental task in advertising and recommendation systems. An accurate CTR prediction can directly improve user experience~\cite{kaasinen2009user} and enhance ads profit~\cite{Wang2020ASO}. 

In a typical industrial dataset, the number of click samples has grown up to hundreds of billion~\cite{Zhao2019AIBoxCP, Xie2020KrakenMC} and keeps increasing on a daily basis. The click-through rate (CTR) prediction task is to predict whether a user will click on the recommended item. It is a fundamental task in advertising and recommendation systems. Since CTR prediction is a time-sensitive task~\cite{Zhao2019AIBoxCP} (\eg, latest topics, hottest videos, and new users' hobbies), it is necessary to shorten the time needed for re-training on a massive dataset to maintain an up-to-date CTR prediction model. In addition, given a constant computing budget, decreasing the training time also reduces the training cost, giving rise to a high return-to-investment ratio.

\begin{figure}[t]
\centering
\includegraphics[width=\columnwidth,trim={0pt 0pt 50pt 0pt},clip]{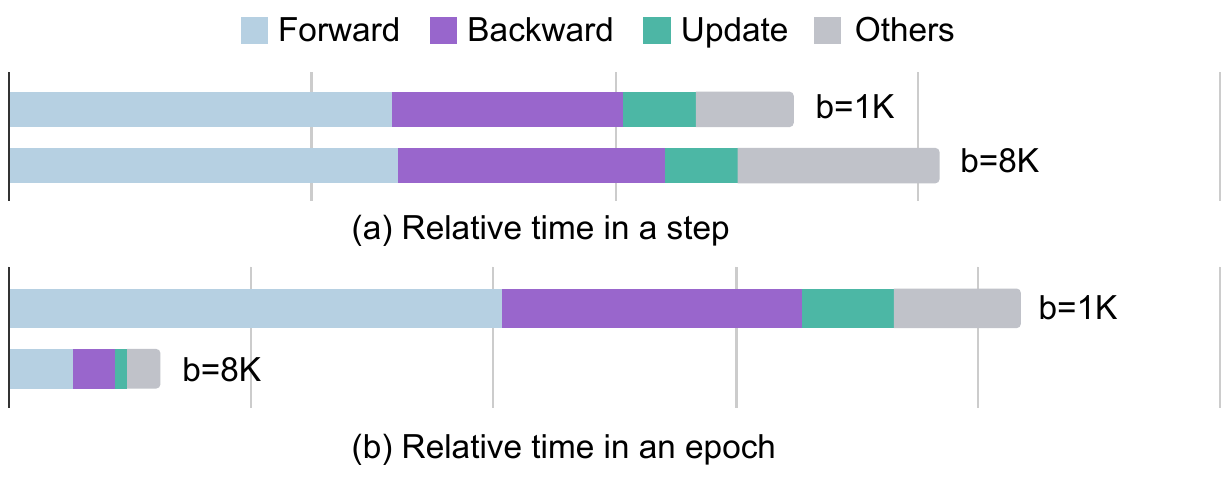}
\caption{Relative time of training DeepFM model on Criteo dataset with one V100 GPU.}
\label{fig:time_per}
\end{figure}

\begin{figure}[t]
  \centering
  \includegraphics[width=.99\linewidth]{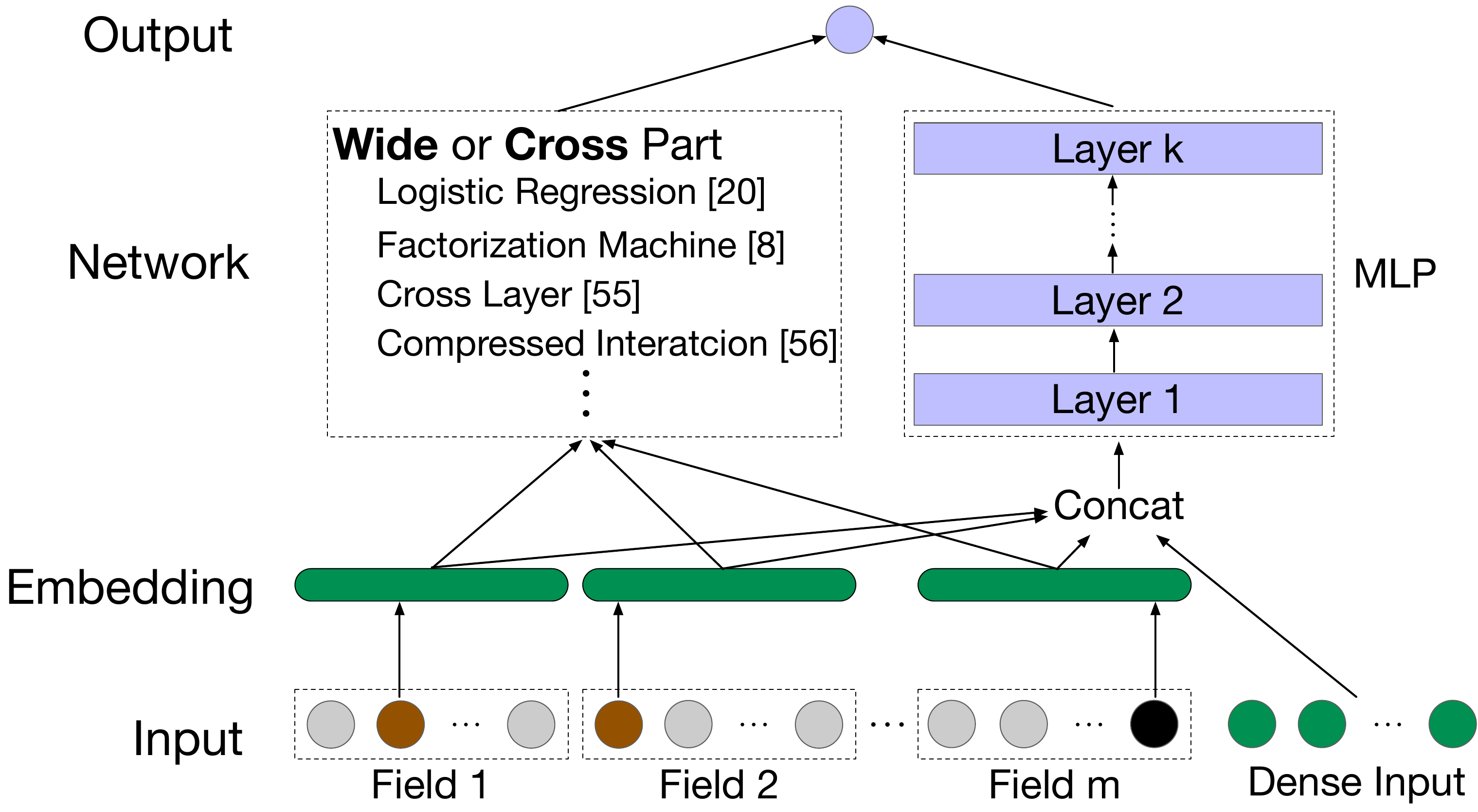}
\caption{A simple illustration of a Wide/Cross-and-Deep style of CTR prediction model. The \textcolor{f3green}{green} data denotes a dense one, while \textcolor{f3brown}{brown} input in a categorical field stands for a selected id.}
\label{fig:f3_right}
\end{figure}

Recent years have witnessed rapid growth in GPU processing ability~\cite{baji2018evolution}.
With the growth of GPU memory and FLOPS, a larger batch size can take better advantage of the parallel processing capability of GPUs. As shown in Figure~\ref{fig:time_per} (a), the time of one forward and backward pass is almost the same when scaling 8 times batch size, indicating GPU with a small batch size is extremely underused. Since the number of training epochs remains the same, large batch training reduces the number of steps and thus significantly shortens the total training time (Figure~\ref{fig:time_per} (b)). In addition, a large batch benefits more in a multi-GPUs setting, where gradients of the large embedding layer need to be exchanged between different GPUs and machines, resulting in high communication costs. To avoid distraction from system optimization in reducing communication costs~\cite{Mudigere2021SoftwareHardwareCF, Zhao2019AIBoxCP, Xie2020KrakenMC}, we focus on designing an accuracy-preserving algorithm for scaling batch size on a single GPU, which can be easily extended for multi-node training.
The challenge of applying large batch training is an accuracy loss when naively increasing the batch size~\cite{He2021LargeScaleDL}, especially considering that
CTR prediction is a very sensitive task and cannot bear the accuracy loss. 
Hyperparameter scaling rules~\cite{Krizhevsky2014OneWT,Goyal2017AccurateLM} and carefully designed optimization methods~\cite{You2017LargeBT,You2020LargeBO} in CV and NLP tasks are not directly suitable for CTR prediction. This is because, in CTR prediction, the inputs are more sparse and frequency-unbalanced, and the embedding layers dominate the parameters of the whole network (\eg, 99.9\%, see Table~\ref{tab:nparams}).
In this paper, we identified the failure reason behind previous scaling rules on CTR prediction and proposed an effective algorithm and scaling rule for large batch training.

\begin{table}[t]
\centering
\caption{Number of parameters for different layers.}
\resizebox{\columnwidth}{!}{
\begin{tabular}{@{}lcccccc@{}}
\toprule
 & \multicolumn{4}{c}{
 \bf Network} & \multicolumn{2}{c}{\bf Embedding (Dataset)} \\ \cmidrule(lr){2-5}\cmidrule(lr){6-7}
{\bf Name} & W\&D & DeepFM & DCN & DCNv2 & Criteo & Avazu \\
{\bf \#Params} & 0.431M & 0.431M & 0.433M & 0.655M & 372M & 104M \\ \bottomrule
\end{tabular}
}
\label{tab:nparams}
\end{table}

In conclusion, our contributions are as follows:
\begin{packed_itemize}
    \item To the best of our knowledge, we are the first to investigate the stability of the training CTR prediction model in very large batch sizes. 
    We attribute the hardship in scaling the batch size to the difference in id frequencies.
    \item With rigorous mathematical analysis, we prove that the learning rate for infrequent features should not be scaled when scaling up the batch size. With CowClip, we can adopt an easy and effective scaling strategy for scaling up the batch size.
    \item We propose an effective optimization method of adaptive Column-wise Clipping (CowClip) to stabilize the training process of the CTR prediction task. We successfully scale up 128 times batch size for four models on two public datasets. In particular, we train the DeepFM model with 72 times speedup and 0.1\% AUC improvement on the Criteo dataset.
\end{packed_itemize}

\begin{figure}[t]
    \centering
    \includegraphics[width=.9\linewidth]{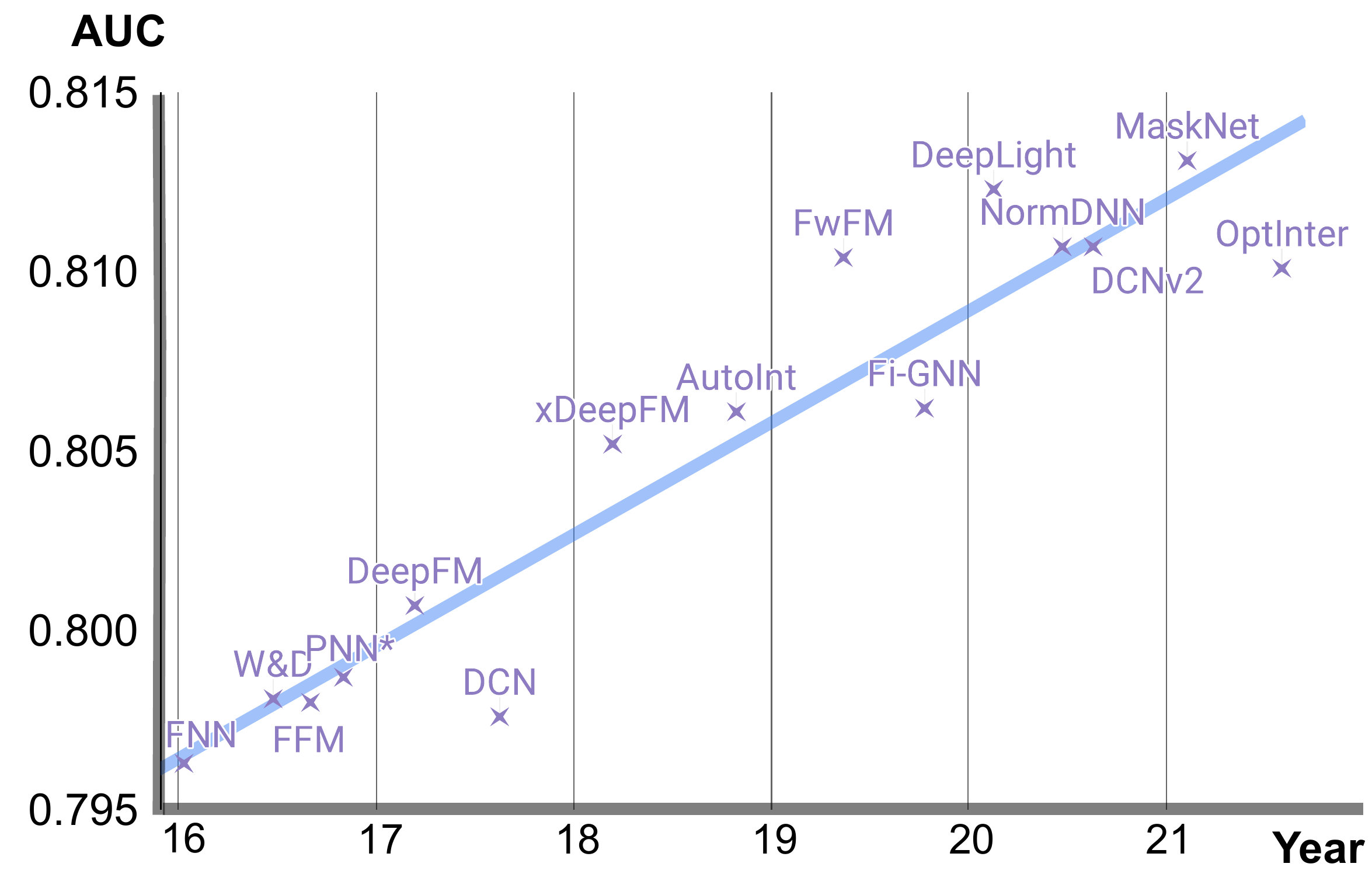}
    \caption{Progress on AUC of CTR prediction models on Criteo dataset in the past six years.}
    \label{fig:f1_left}
\end{figure}

\section{Related Work}
\label{sec:rw}

\paragraph{Embeddings of CTR Prediction model}
\label{app:input}
The input of the CTR prediction model is high-dimensional, sparse, and frequent-unbalanced. As we will discuss the frequency in the Section~3, we focus on the fact that the input feature space for CTR prediction is high-dimensional and sparse, which is an essential difference between the CTR prediction model and other deep learning models.

A typical industrial CTR prediction model~\cite{Zhao2019AIBoxCP, Xie2020KrakenMC, Zhou2019DeepIE} has a high-dimensional input space with $10^8$ to $10^{12}$ dimensions after one-hot encoding of the categorical features. At the same time, a single clicking log may contain only hundreds of non-zero entries. As a result, when we create the embedding for each feature, the whole embedding layer can be extremely large, and the parameters of the CTR prediction model are dominated (\eg, 99.9\%) by the embedding part instead of the deep network part~\cite{Miao2021HETSO,Ginart2021MixedDE}. Table~\ref{tab:nparams} shows the case under our experimental setting. 

As the number of parameters in the embedding layer overwhelms the one of the dense networks, the difficulty of large batch optimization lies in the embedding layers. This paper focuses on addressing the training instability caused by the properties of embedding layers in the CTR prediction model. No matter how the dense part, \eg, MLP, LSTM~\cite{chen2021improved}, Transformer~\cite{chen2019behavior}, changes, the training instability caused by the embedding part still exists.

\paragraph{CTR prediction network}

A thread of work started from~\cite{Cheng2016WideD,Wang2017DeepC} occupies a majority of the above networks. They focused on designing a two-stream network, as shown in Figure~\ref{fig:f1_right}. Following W\&D model~\cite{Cheng2016WideD}, there are many designs on the wide/cross-stream. The details of DeepFM~\cite{Guo2018DeepFMAE}, W\&D, DCN~\cite{Wang2017DeepC}, and DCN-v2~\cite{Wang2021DCNVI} used in our experiments are presented in the Appendix~\ref{app:wdm}.

\begin{table}[t]
\centering
\caption{AUC (\%) changes at different batch sizes on Criteo with DeepFM and a modified version. Previous scaling rules fail on Criteo but work for a revised version.}
{\small
\begin{tabular}{@{}cccc@{}}
\toprule
 & No Scale & Sqrt Scale & Linear Scale \\ \midrule
 & \multicolumn{3}{c}{\bf Criteo} \\ \cmidrule(lr){2-4}
1k & 80.76 & 80.76 & 80.76  \\
2k & --0.15 & --0.01 & --0.01  \\
4k & --1.35 & --0.06 & --0.11 \\
8k & --3.21 & --0.21 & --0.20  \\ \midrule
 & \multicolumn{3}{c}{\bf Criteo (Top 3 frequent ids)} \\ \cmidrule(lr){2-4}
1k & 74.97 & 74.97 & 74.97 \\
2k & --0.10 & --0.01 & +0.04 \\
4k & --0.20 & --0.02 & --0.02 \\
8k & --0.28 & --0.01 & --0.01 \\ \bottomrule
\end{tabular}
}
\label{tab:fail}
\end{table}

\paragraph{Large batch training methods.} 
To preserve the performance of deep models at a large batch size, we need a good scaling rule and a stable optimization strategy. The scaling rule tells us how to scale the hyperparameters when scaling up the batch size. The two most important hyperparameters when scaling the batch size are learning rate and regularization weight. Based on different assumptions, linear scaling~\cite{Goyal2017AccurateLM} and square root scaling~\cite{Krizhevsky2014OneWT,Hoffer2017TrainLG} are the two most common scaling rules in the deep learning community. Besides, optimization strategies such as warmup~\cite{Gotmare2019ACL} and gradient clipping~\cite{Zhang2020WhyGC} can help stabilize the large batch training process. LARS~\cite{You2017LargeBT} and LAMB~\cite{You2020LargeBO} are two optimizers designed for large batch training, which adopt different adaptive learning rates for each layer. Although they achieve good results in CV and NLP tasks, they are ineffective in the CTR prediction task because it is unnecessary to use a layer-wise optimizer with a shallow network (\eg, three or four layers). This paper re-designs the scaling rule and optimization strategy for the embedding layer, which can successfully scale up the batch size for CTR prediction.

\paragraph{} Additional related work can be found in Appendix~\ref{app:wdm}, including sensitiveness of CTR prediction and works utilizing different frequencies.

\section{Method}

In CTR prediction, we have the training dataset $\mathcal{D}=\{\bm{x}_i,y_i\}_{i=1}^N$, where $y\in\{0,1\}$ denotes whether the user clicked or not. The $\bm{x}$ contains information about the user, the product, and the interaction, which can be categorical or continuous. The categorical field is one-hot encoded to be a vector $\bm{x}_{i}^{\text{f}_j}$ of $d_{\text{f}_j}$ length, where $d_{\text{f}_j}$ is the number of possible values (ids) in this field. 
To represent the frequency of each id, we denote the $k$-th id in field $j$ as $\text{id}_k^{\text{f}_j}$. The frequency and occurrence probability of the id is:
% \vspace{-5pt}
\begin{align*}
    \occur(\text{id}_k^{\text{f}_j}) &=\sum_{i=1}^N\delta(\bm{x}_{i}^{\text{f}_j}[k]=1),\\
    \freq &=\frac{\texttt{count}(\text{id}_k^{\text{f}_j})}{N},
\end{align*}
where $\delta(\cdot)$ equals $1$ if the boolean condition holds and $0$ otherwise.

\begin{figure}[t]
    \centering
    \includegraphics[width=.75\columnwidth]{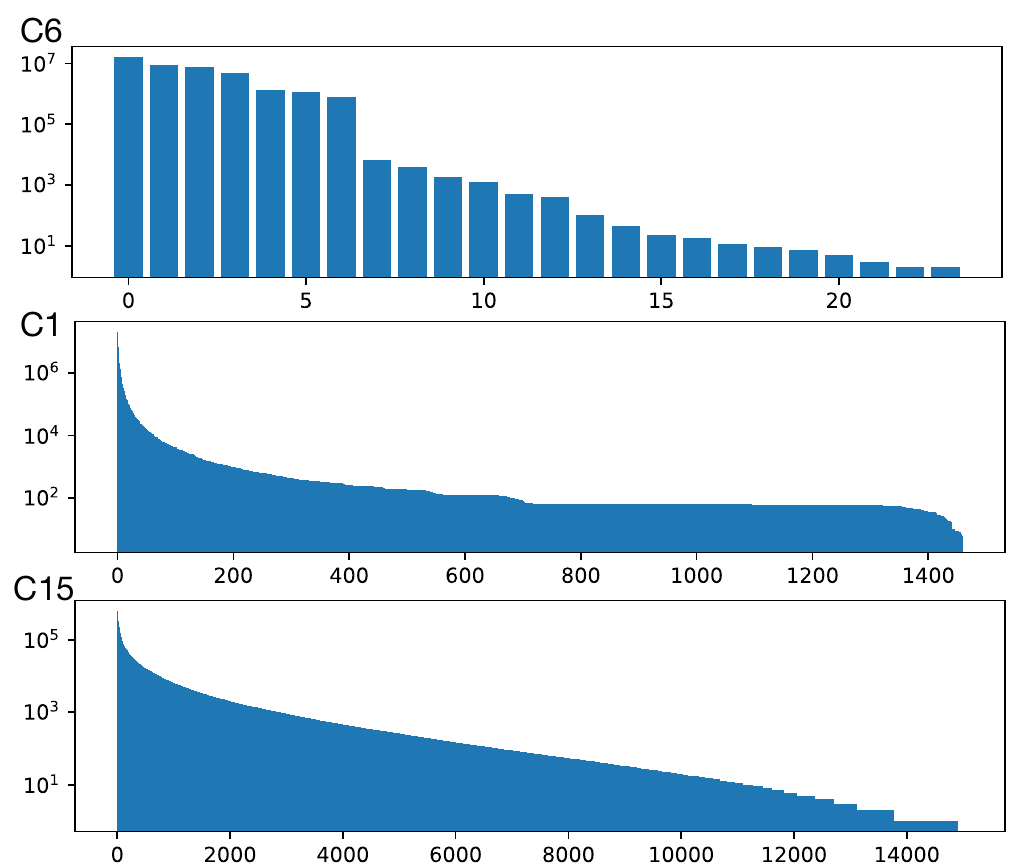}
    \caption{Distribution of different ids in three fields of the Criteo dataset. The y-axis is in logarithm scale. The total number of samples is $4.13\times 10^{7}$.}
\label{fig:f3_left}
\end{figure}

Given the predicting network $f$, the prediction is made from $f(\bm{x})$. The network and embeddings weights are denoted as $w$, and the training loss is $L$. This paper focuses on the Wide/Cross-and-Deep kind of CTR prediction model, as briefly described in Figure~\ref{fig:f3_right}, one of the state-of-the-art networks in CTR prediction~\cite{Wang2021DCNVI,Zhang2019FATDeepFFMFA}.

In training the network, we use a batch size of $b=|B|$, where $B$ is a specific batch. The learning rate and L2-regularization weight are denoted as $\eta$ and $\lambda$. The total number of steps in an epoch is $\frac{N}{b}$.

\subsection{Failure cause of traditional scaling rules}
\label{subsec:fail}

When training a neural network, at step $t$, an optimizer $\texttt{Opt}(\cdot)$ takes in the weights and gradients, and output the updated weights. With the L2-regularization, the update process can be formulated as:
\begin{align*}
    \bm{g}_t &= \sum_{x\in B_t}\nabla L(w,x) + \frac{\lambda}{2}\cdot\|w\|^2_2\\
    w_{t+1}&=\eta\cdot\texttt{Opt}(w_t, \bm{g}_t).
\end{align*}
When changing the batch size, the hyper-parameter learning rate $\eta$ and L2-regularization weight $\lambda$ should be adjusted for maintaining the same performance as the original batch size.

Square root scaling~\cite{Krizhevsky2014OneWT,Hoffer2017TrainLG} and linear scaling~\cite{Goyal2017AccurateLM} are two widely used scaling rules in deep learning. The motivation for sqrt scaling is to keep the covariance matrix of the parameters update the same, while for linear scaling, the motivation is to keep the update in a large batch equal to updates from $s$ small batches when scaling $s$ times the batch size (details in Appendix~\ref{app:tradderi}). The two scaling rules have been shown effective in CV and NLP tasks, and they are shown as follows:
\begin{scalingrule}[Sqrt Scaling]
    When scaling batch size from $b$ to $s\cdot b$, do as follows:
    \[\eta\rightarrow \sqrt{s}\cdot\eta,\quad \lambda\rightarrow \sqrt{s}\cdot\lambda\]
\end{scalingrule}
\begin{scalingrule}[Linear Scaling]
    When scaling batch size from $b$ to $s\cdot b$, do as follows:
    \[\eta\rightarrow s\cdot\eta,\quad \lambda\rightarrow \lambda\]
\end{scalingrule}
Our first attempt at large batch training of the CTR prediction model is to apply the above classic scaling rules: no scaling, linear scaling~\cite{Goyal2017AccurateLM}, and square root scaling~\cite{Krizhevsky2014OneWT}. 
However, as seen in experiments on the Criteo dataset with DeepFM model in Table~\ref{tab:fail} left, the above rules fail in a CTR prediction model. We claim that the reason for the failure lies in the different frequencies of ids.

The product id of a popular item or ids in fields with a few options (\eg, male and female in gender field) are frequent, while the id of an inactive user seldom appears. In Figure~\ref{fig:f3_left}, we visualize the distribution of different ids' frequencies in three fields. The exponential distribution reveals different frequencies among different ids. For the dense weights (\eg, kernel weights), their gradients appear for each sample while embedding does not have gradients if the corresponding ids do not show up. In CTR prediction, the embedding layers dominate the parameters of the whole network, and different occurrences of gradients make a great difference from other deep neural networks.

First, we empirically verify our claim by the following experiment.
We keep the top three frequent ids in each field and label the rest as a fourth id. In this way, all four ids are very frequent and variations in frequencies are ablated in this modified version of Criteo. As shown in Table~\ref{tab:fail} right, both scaling rules successfully apply to the modified dataset, which means the traditional scaling rule does not work in CTR prediction due to the presence of infrequent ids.

Next, we provide the theoretical analysis for the failure of sqrt and linear scaling. Different frequencies lead to varying occurrences of ids in batches. Only when the id appears in the batch can the corresponding embedding be updated. They only occur in a small fraction of batches for ids with a low frequency.
Suppose we draw the training samples with replacement from the dataset, the probability of an id $\text{id}_k^{\text{f}_j}$ in the batch $B$ is:
\begin{equation*}
    \prob(\text{id}_k^{\text{f}_j} \in B)=1-(1-\freq)^b.
\end{equation*}
For frequent ids, and also dense weights whose frequency rate is $1$, we have $(1-\freq)^b\approx 0$; while for the infrequent ids, when $p\ll\frac{1}{B}$, we can use binomial approximation and obtain:
\begin{equation}
\label{eq:case}
    \prob(\text{id}_k^{\text{f}_j} \in B) \approx \begin{cases}
    1 & \text{id}_k^{\text{f}_j}\text{ is frequent}\\
    b\cdot \freq & \text{id}_k^{\text{f}_j}\text{ is infrequent}
\end{cases}.
\end{equation}
Now, reconsider the linear scaling motivation for an id's embedding $w$. 
Denote the weight update as $\Delta w=w_t-w_{t+1}$. Consider the expected update in a large batch $B'=\bigcup_{i=1}^sB_i$ with $b'=|B'|=s\cdot b$, we have 
\begin{align*}
    \E[\Delta w] &= \E[\eta'\cdot\delta(\text{id}_k^{\text{f}_j} \in B')\cdot\frac{1}{b'}\sum_{x\in B'}\nabla L(w, x)] \\
    &=\eta'\cdot \prob(\text{id}_k^{\text{f}_j} \in B')\cdot \E[\nabla L(w, x)].
\end{align*}
With the assumption that $\E[\nabla L(w_{i}, x)]\approx \E[\nabla L(w, x)]$, the expected update in small batches $B_i$ is:
\begin{align*}
    \E[\Delta w] &= \E[\eta\cdot\sum_{i=1}^{s}\delta(\text{id}_k^{\text{f}_j} \in B_i)\cdot\frac{1}{b}\sum_{x\in B_{i}}\nabla L(w_{i}, x)]\\ &\approx\eta\cdot s\cdot \prob(\text{id}_k^{\text{f}_j} \in B)\cdot \E[\nabla L(w, x)].
\end{align*}
For dense weight or embeddings of frequent id, the term $\prob(\text{id}_k^{\text{f}_j}\in B)$ equals $1$, making no difference to the original linear scaling rule. However, with an infrequent id, it shows that the new scaling strategy should be using the same learning rate when scaling the batch size due to the following fact for infrequent ids:
\begin{equation*}
    \prob(\text{id}_k^{\text{f}_j}\in B') \approx s\cdot \prob(\text{id}_k^{\text{f}_j}\in B).
\end{equation*}
A similar discussion based on sqrt scaling motivation (see Appendix~\ref{app:sqrt}) shows that under a very strong assumption can we obtain the same conclusion. However, without the assumption, we cannot even choose hyperparameters maintaining the same covariance matrix after scaling the batch size.

When using a relatively small batch size
we find most ids satisfied $p<\frac{1}{B}$. Thus, we propose to use no scaling on the whole embedding layers, which suits infrequent ids. 
In addition, a smaller learning rate for layers at the bottom leads to a smooth learning process. Experiments show this scaling rule leads to a better result.

After the discussion of learning rate scaling, now let's turn to the L2-regularization weight $\lambda$. In CTR prediction, an unsuitable $\lambda$ can easily lead to overfitting.
For the scaling of $\lambda$, we first consider the embedding vector $w$ of $\text{id}_k^{\text{f}_j}$, the expected gradient of which in a batch is:
\begin{align}
    \E[\bm{g}] &= \frac{1}{b}\E[\delta(\text{id}_k^{\text{f}_j}\in B)\sum_{x\in B}\nabla L(w, x)] \nonumber \\
    &= \prob(\text{id}_k^{\text{f}_j}\in B)\cdot \E[\nabla L(w, x)].\label{eq:pe}
\end{align}
The term $\prob(\text{id}_k^{\text{f}_j}\in B)$ still has no effect with dense weight and embeddings of frequent ids as the probability equals to $1$. However, for the infrequent ids, there is a scaling multiplier before the expectation of the gradient as some ids may not appear in a certain batch.
When using an adaptive optimizer such as Adam, this scaling multiplier results in a different behaviour, which is equivalent to adjusting the L2-regularization weight $\lambda$ as follows (see Appendix~\ref{app:adam} for the proof):
\begin{equation*}
    \frac{\lambda}{\prob(\text{id}_k^{\text{f}_j}\in B)} = \frac{\lambda}{b\cdot\freq}.
\end{equation*}
Thus, to maintain the same L2-regularization strength, we scale up the $\lambda$ by $n$. Combined with the learning rate scaling rule, we have the following one.
\begin{scalingrule}[CowClip Scaling]
    \label{sr4}
    When scaling batch size from $b$ to $s\cdot b$, use sqrt scaling for the dense weights, and do as follows for embeddings:
    \[\eta_e\rightarrow \eta_e,\quad \lambda\rightarrow s\cdot\lambda\]
\end{scalingrule}

% The norm of weights and gradients at 1st iteration
% fig: embedding gradient norm curve
% step = 1000
\begin{figure}[t]
    \centering
    \includegraphics[width=0.75\columnwidth,trim={20pt 20pt 20pt 40pt},clip]{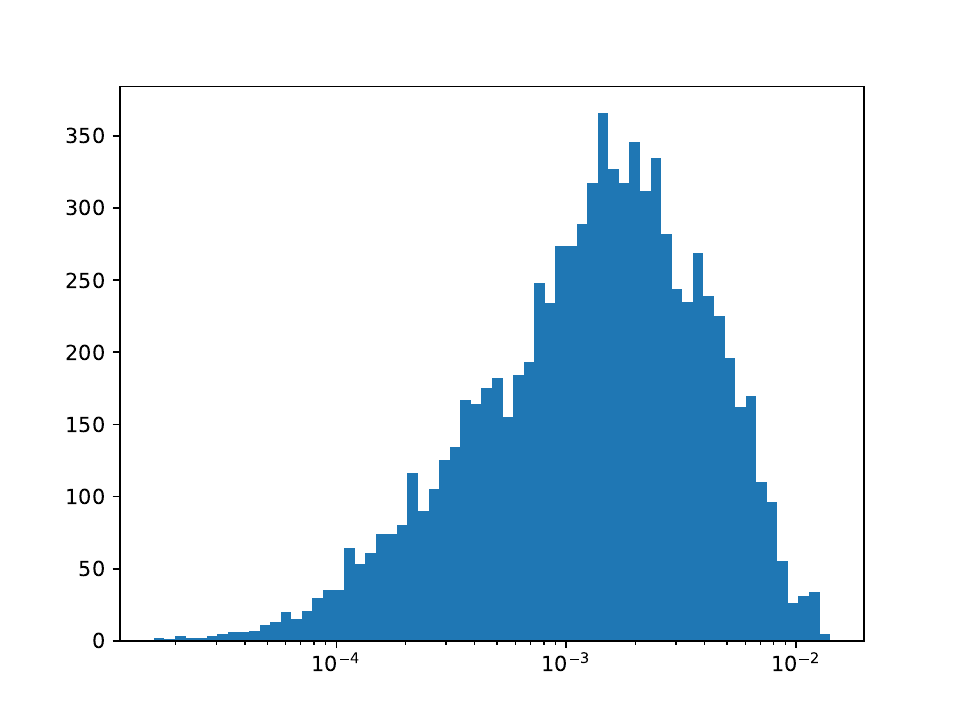}
    \caption{L2-norm distribution of different columns gradients at 1000th step of DeepFM on Criteo dataset. Only columns with existing ids in the batch are shown. The x-axis is the L2 norm value and y-axis is the count of columns.}
    \label{fig:s1000gnorm}
\end{figure}

However, we find directly applying the above rule leads to overfitting due to $s$ times less application of L2-regularization when scaling up batch size by $s$ times. If no additional regularization technique is introduced, L2-regularization should be strengthened further with a large batch size. In the case of an SGD optimizer, we have (details in Appendix~\ref{app:tradderi}):
\begin{equation*}
    \eta'\lambda'\approx s\eta\lambda.
\end{equation*}
Hence, we need to further scale up the $\lambda$ by $s$ times when the learning rate is unchanged. Although the behavior of adaptive optimizers such as Adam is different from SGD, we find a larger $\lambda$ prevents overfitting. Thus, we have the following scaling rule which can scale up the batch size to 4K without additional optimization strategy.  
\begin{scalingrule}[$n^2$--$\lambda$ Scaling]
\label{sr3}
    When scaling batch size from $b$ to $s\cdot b$, use sqrt scaling for the dense weights, and do as follows for embeddings:
    \[\eta_e\rightarrow \eta_e,\quad \lambda_e\rightarrow s^2\cdot\lambda_e\]
\end{scalingrule}

\subsection{CowClip algorithm}

Although the Scaling Rule~\ref{sr3} helps us scale to 4 times the original batch size, it fails on a larger batch size. The challenge in choosing the proper learning rate $\eta$ and L2-weight $\lambda$ mentioned above impairs the performance for  larger batch sizes. 
To enable large batch training, the gradient norm clipping~\cite{Zhang2020WhyGC} can smooth the process of training and alleviate the sensitiveness of hyperparameters. Given a clip threshold $\texttt{clip\_t}$, gradient norm clipping does follows:
\begin{equation*}
    \bm{g}\rightarrow \min\{1, \frac{\texttt{clip\_t}}{\|\bm{g}\|}\}\cdot \bm{g}
\end{equation*}
Gradient norm clipping smoothes the training process by reducing the norm of a large gradient greater than a threshold. However, it is hard to choose an appropriate threshold for clipping the norm. Besides, as one column of the embedding matrix represents the embedding vector for an id, Figure~\ref{fig:s1000gnorm} shows that the magnitude of gradients for different columns varies. We denote the column for an id as $w[\text{id}_k^{\text{f}_j}]$. Clipping on the whole embeddings whose gradient norm is dominated by gradients of columns with large gradients impairs the ones with normal but smaller gradients. In addition, according to Equation (\ref{eq:pe}), since we want to clip on $1\cdot\nabla L(w, x)$, the different frequencies of ids lead to the scaler of $\prob(\text{id}_k^{\text{f}_j}\in B)$ on the expected gradients. 

To tackle the above problems, inspired by LAMB optimizer~\cite{You2020LargeBO}, which normalizes the norm of gradients of each kernel to be proportional to the norm of the kernel weight, we relate the clip threshold with the norm of id embedding vectors. The difference between our clipping method and the gradient norm clipping is three-fold: First, every id embedding vector has a unique clipping threshold for more flexible clipping. Second, the clipping threshold is multiplied by the occurrence number of the id to make sure the bound is based on $1\cdot\nabla L(w, x)$. Last but not the least, the clipping threshold is calculated by the norm of the id vector in consideration of different magnitudes:
\begin{equation*}
    \texttt{clip}(\texttt{id}_k^{\text{f}_j})=\text{cnt}(\text{id}_k^{\text{f}_j})\cdot\max\{r\cdot \|w_t^e[\text{id}_k^{\text{f}_j}]\|\,, \,\,\zeta\}
\end{equation*}
where $\texttt{cnt}(\text{id}_k^{\text{f}_j})$ is the number of occurence of the id in a batch.

As the weights grow larger in the training process, the benefit of a threshold proportional to the norm of the weight is that the clipping value adaptively grows with the network. As some infrequent id embedding vectors become too small due to the continual application of L2-regularization with no id occurrence in steps, we restrict the clipping norm by a lower-bound $\zeta$ to avoid a too strong clipping.

\begin{algorithm}[t]
\caption{Adaptive Column-wise Clipping(CowClip)}
\begin{algorithmic}[1]
    \REQUIRE CowClip coefficient $r$ and lower-bound $\zeta$, number of steps $T$, batch size $b$, learning rate for dense and embedding $\eta,\eta_e$, optimizer $\texttt{Opt}(\cdot)$
    % \ENSURE network weights $w$, embedding weights $w^e$
    \FOR{$t \gets 1$ to $T$}
        \STATE Draw $b$ samples $B$ from $\mathcal{D}$
        \STATE $\bm{g}_t,\bm{g}^e_t \gets \frac{1}{b}\sum_{x\in B}\nabla L(x,w_t,w_t^e)$
        \STATE $w_{t+1} \gets \eta\cdot\texttt{Opt}(w_t, \bm{g}_t)$ \COMMENT{Update dense weights}
        \FOR{each field and each column in the field}
            \STATE $n_{\bm{g}} \gets \|\bm{g}_t^e[\text{id}_k^{\text{f}_j}]\|$
            \STATE $\texttt{cnt} \gets |\{x\in B|\text{id}_k^{\text{f}_j}\in x\}|$ 
            \STATE \COMMENT{Calculate the number of occurrence $\texttt{cnt}$}
            \STATE $\texttt{clip\_t}\gets \texttt{cnt}\cdot\max\{r\cdot \|w_t^e[\text{id}_k^{\text{f}_j}]\|\,, \,\,\zeta\}$
            \STATE \COMMENT{Clip norm threshold}
            \STATE $\bm{g}_c \gets \min\{1,\frac{\texttt{clip\_t}}{n_{\bm{g}}}\}\cdot \bm{g}_t^e[\text{id}_k^{\text{f}_j}]$ 
            \STATE \COMMENT{Gradient clipping}
            \STATE $w_t^e[\text{id}_k^{\text{f}_j}] \gets \eta_e\cdot\texttt{Opt}(w_t^e[\text{id}_k^{\text{f}_j}], \bm{g}_c)$ 
            \STATE \COMMENT{Update the id embedding}
        \ENDFOR
    \ENDFOR
\end{algorithmic}
\label{alg}
\end{algorithm}

\begin{table*}[t]
\centering
\caption{Performance comparison between Cowclip and previous scaling methods with different batch size.}
\label{tab:lb_all}
% \resizebox{\columnwidth}{!}{
\begin{tabular}{@{}ccccccc@{}}
\toprule
 & \multicolumn{2}{c}{\bf 1K} & \multicolumn{2}{c}{\bf 8K} & \multicolumn{2}{c}{\bf 128K} \\ \midrule
 & Prev. best & CowClip & Prev. best & CowClip & Prev. best & Cowclip \\ \midrule
Criteo & 80.76 & 80.86 & 80.55 & 80.97 & -- & 80.90 \\
Criteo-seq & 80.48 & 80.50 & 80.03 & 80.50 & -- & 80.49 \\
Avazu & 78.84 & 78.83 & 76.69 & 79.06 & -- & 78.80 \\ \bottomrule
\end{tabular}
% }
\end{table*}

\begin{table*}[t]
\caption{Performance of different scaling methods on Criteo dataset from 1K to 8K on DeepFM.}
\centering
\resizebox{0.8\textwidth}{!}{%
\begin{tabular}{@{}lcccccccc@{}}
\toprule
 & \multicolumn{2}{c}{\bf 1K (1024)} & \multicolumn{2}{c}{\bf 2K (2048)} & \multicolumn{2}{c}{\bf 4K (4096)} & \multicolumn{2}{c}{\bf 8K (8192)} \\\cmidrule(lr){2-3}\cmidrule(lr){4-5}\cmidrule(lr){6-7}\cmidrule(lr){8-9}
 & AUC (\%) & LogLoss & AUC (\%) & LogLoss & AUC (\%) & LogLoss & AUC (\%) & LogLoss \\ \midrule
No Scaling & 80.76 & 0.4438 & 80.66 & 0.4456 & 80.48 & 0.4518 & 80.31 & 0.4530 \\
Sqrt Scaling & 80.76 & 0.4438 & 80.71 & 0.4430 & 80.59 & 0.4450 & 80.28 & 0.4582 \\
Sqrt Scaling$^*$ & 80.76 & 0.4438 & 80.75 & 0.4444 & 80.69  & 0.4449  & 80.55 & 0.4547\\
LR Scaling & 80.76 & 0.4438 & 80.77 & 0.4434 & 80.65 & 0.4434 & 80.46 & 0.4542 \\
$n^2$--$\lambda$ Scaling (Ours) & 80.76 & 0.4438 & 80.86 & 0.4432 & 80.90 & 0.4426 & 80.73 & 0.4441 \\
CowClip (Ours) & 80.86 & 0.4430 & 80.93 & 0.4427 & 80.97 & 0.4422 & 80.97 & 0.4425 \\ \bottomrule
\end{tabular}%
}
\label{tab:lb_criteo_8k}
% \vspace{-5pt}
\end{table*}

The network training with CowClip is summarized in the Algorithm~\ref{alg}. In practice, tensor multiplication instead of for-loop is adopted for less computational overhead. 
% The implementation can be found at  \href{https://github.com/zhengzangw/LargeBatchCTR}{this https URL}. 
Since CowClip stabilizes the training process, it is possible to use the CowClip scaling~\ref{sr4} rule to 128$\times$ batch size, leaving $\eta_e$ unchanged and linear scaling the $\lambda$. We give a proof sketch on the convergence of CowClip method in Appendix~\ref{sec:proof}. Our large batch training framework contains the CowClip gradient clipping and scaling strategy.

\begin{table*}[t]
\caption{Performance of CowClip methods on Criteo dataset from 1K to 128K on four models.}
\centering
\resizebox{0.85\textwidth}{!}{%
\begin{tabular}{@{}clccccccccc@{}}
\toprule
 &  & Baseline & 1K & 2K & 4K & 8K & 16K & 32K & 64K & 128K \\ \midrule
\multirow{2}{*}{DeepFM~\cite{Guo2018DeepFMAE}} & AUC (\%) & 80.76 & 80.86 & 80.93 & 80.97 & 80.97 & 80.94 & 80.95 & 80.96 & 80.90 \\
 & Logloss & 0.4438 & 0.4430 & 0.4427 & 0.4422 & 0.4425 & 0.4424 & 0.4423 & 0.4429 & 0.4430 \\ \midrule
\multirow{2}{*}{W\&D~\cite{Cheng2016WideD}} & AUC (\%) & 80.75 & 80.86 & 80.94 & 80.96 & 80.96 & 80.95 & 80.94 & 80.96 & 80.89 \\
 & Logloss & 0.4439 & 0.4430 & 0.4424 & 0.4422 & 0.4425 & 0.4422 & 0.4428 & 0.4429 & 0.4434 \\ \midrule
\multirow{2}{*}{DCN~\cite{Wang2017DeepC}} & AUC (\%) & 80.76 & 80.86 & 80.93 & 80.96 & 80.97 & 80.98 & 80.95 & 80.99 & 80.91 \\
 & Logloss & 0.4438 & 0.4429 & 0.4424 & 0.4422 & 0.4428 & 0.4419 & 0.4426 & 0.4426 & 0.4428 \\ \midrule
\multirow{2}{*}{DCN v2~\cite{Wang2021DCNVI}} & AUC (\%) & 80.78 & 80.87 & 80.94  & 80.97 & 80.98 & 80.97 & 80.95 & 80.97 & 80.89 \\
 & Logloss & 0.4437 & 0.4429 & 0.4425 & 0.4422 & 0.4423 & 0.4420 & 0.4424 & 0.4427 & 0.4427 \\ \bottomrule
\end{tabular}%
}
\label{tab:lb_criteo_128k}
% \vspace{-5pt}
\end{table*}

\begin{table}[t]
\caption{The training time of different methods on Criteo dataset. Last four are trained with CowClip.}
\centering
\resizebox{\columnwidth}{!}{%
\begin{threeparttable}
\begin{tabular}{@{}lcccccccccc@{}}
\toprule
 &  &  & \multicolumn{8}{c}{\bf Time (minutes)} \\ \cmidrule(lr){3-11}
 & AUC (\%) & Logloss & 1K & 2K & 4K & 8K & 16K & 32K & 64K & 128K \\ \midrule
XDL & 80.2 & 0.452 & 196 & 179$^\dagger$ & 160$^\ddagger$ & -- & -- & -- & -- & -- \\
FAE & 80.2 & 0.452 & 122 & 116$^\dagger$ & 104$^\ddagger$ & -- & -- & -- & -- & -- \\
DLRM & 79.8 & 0.456 & 196 & 133$^\dagger$ & 76$^\ddagger$ & -- & -- & -- & -- & -- \\
Hotline & 79.8 & 0.456 & 53 & 45$^\dagger$ & 39$^\ddagger$ & -- & -- & -- & -- & -- \\ \midrule
DeepFM & 80.87 & 0.4428 & 768 & 390 & 204 & 102 & 48 & 27 & 15 & 9 \\
W\&D & 80.86 & 0.4430 & 768 & 390 & 204 & 102 & 48 & 27 & 15 & 10 \\
DCN & 80.86 & 0.4429 & 768 & 390 & 204 & 102 & 48 & 28 & 17 & 11 \\
DCN v2 & 80.87 & 0.4429 & 822 & 408 & 210 & 108 & 60 & 40 & 34 & 30 \\ \midrule
\multicolumn{2}{l}{\hspace{-6pt}Speedup (DeepFM)} & & 1$\times$ & 1.96$\times$ & 3.76$\times$ & 7.52$\times$ & 16.00$\times$ & 28.44$\times$ & 51.2$\times$ & 76.8$\times$ \\ \bottomrule
\end{tabular}%
\begin{tablenotes}
    \footnotesize
    \item[$\dagger$] Trained with 2 GPUs $^\ddagger$ Trained with 4 GPUs.
\end{tablenotes}
\end{threeparttable}
}
\label{tab:lb_criteo_time}
\end{table}

\section{Experiment}
\label{sec:exp}

\subsection{Experimental setting}

\paragraph{Datasets.} We evaluate our algorithms on the following public datasets which are widely adopted by the community~\cite{Cheng2016WideD,Li2019FiGNNMF,Deng2021DeepLightDL,Wang2021DCNVI,Miao2021HETSO}. \textbf{Criteo}~\cite{criteo} is a real-world CTR prediction dataset. It collects 45M records on ad display information, and the corresponding user clicks feedback. There are 13 continuous fields and 26 categorical fields, which are all anonymized to protect users' privacy. Following~\cite{Guo2018DeepFMAE,Zhang2019FATDeepFFMFA}, the data is split into training and test sets by 90\%:10\%. \textbf{Criteo-seq} is a sequential learning setting of Criteo dataset. The first six days' data are used for training and the last day's data for testing. This setting evaluates the performance of the algorithm in a sequential learning setting. \textbf{Avazu}~\cite{avazu} is another ad click-through dataset containing 32M training samples. It has 24 anonymous categorical fields. According to~\cite{Zhang2019FATDeepFFMFA}, we split the dataset into training and test sets by 80\%:20\%.

{\bf Implementation details.} We use two popular metrics~\cite{Mattson2020MLPerfTB} in CTR prediction: AUC (Area Under ROC) and Logloss (Logistic loss). Our implementation is based on Tensorflow~\cite{tensorflow2015-whitepaper} and DeepCTR~\cite{shen2017deepctr} framework. The experiments are conducted on one Tesla V100 GPU. We use Adam~\cite{Kingma2015AdamAM} optimizer and an L2-regularization on embedding layers. The base learning rate and L2-regularization weight on batch size 1024 are $10^{-4}$ and $10^{-5}$. Scaling rules are performed based on the 1024 batch size. For CowClip, we use $r=1$ and tune $\zeta\in\{10^{-5},10^{-4}\}$ due to a different initialization weight norm. We also use learning rate warmup~\cite{Gotmare2019ACL} and larger initialization weights. More discussion on hyperparameter choice and techniques can be found in the Appendix~\ref{app:impl}. We run our experiments with three random seeds, and the standard deviation among all experiments for AUC is less than 0.012\%.

{\bf Baselines.} Four CTR prediction models are considered in our experiments: Wide-and-Deep Network (\textbf{W\&D})~\cite{Cheng2016WideD}, \textbf{DeepFM}~\cite{Guo2018DeepFMAE}, Deep-and-Cross Network (\textbf{DCN})~\cite{Wang2017DeepC}, \textbf{DCN v2}~\cite{Wang2021DCNVI}. The architectures of these networks are detailed in Appendix~\ref{app:wdm}). For the scaling strategy, \textbf{No Scaling} means we use the same hyper-parameters as the ones in batch size 1K. \textbf{Sqrt Scaling} and \textbf{LR Scaling} are described in Section~3. \textbf{Sqrt Scaling$^*$} is a variant version of Sqrt Scaling used in \cite{Guo2018DeepFMAE}, which does not scale up the L2-regularization. For batch size from 1K to 8K, we also do a grid search on learning rate and the weight decay, but it turns out no simple combination yields better results than the above scaling methods. \textbf{DLRM}~\cite{Naumov2019DeepLR} uses model parallelism on the embedding table to accelerate the training. \textbf{XDL}~\cite{Adnan2021AcceleratingRS} is a highly optimized implementation of the above model. \textbf{FAE}~\cite{Adnan2021AcceleratingRS} takes the frequency of embeddings into consideration as well and uses a hot-embedding aware data layout in the memory. \textbf{Hotline}~\cite{Adnan2021AcceleratingID} better organizes the frequent and infrequent embeddings in the GPU and main memory.  \textbf{CowClip} denotes training with the CowClip method and the CowClip scaling rule~\ref{sr4}.

\subsection{Large batch training results}

First, as shown in Table~\ref{tab:lb_all}, previous scaling strategy fails to maintain the performance at batch size 8K, and fails to converge at batch size 128K. In contrast, CowClip methods can achieve a better accuracy at 8K batch size and almost no performance loss at batch size 128K, which shows the CowClip method successfully stablizes the training processes.

Then, we compare different scaling strategies on the DeepFM model. The results on Criteo dataset are presented in Table~\ref{tab:lb_criteo_8k}, and the results on Criteo-seq and Avazu are attached in Appendix. As we can see, traditional scaling rules fail to meet the AUC requirement with a large gap when the batch size grows up to 4K. This is consistent with results in~\cite{Guo2018DeepFMAE}, where results with 4K batch size are worse than those with 1K. With $n^2$--$\lambda$ Scaling rule~\ref{sr3}, it can scale batch size to 4K but fails with 8K. When we successfully scale the batch size, there is a performance gain in the AUC, which is also observed in~\cite{Zhu2021OpenBF}. For our CowClip algorithm, it outperforms the original optimization method by $0.1\%$ AUC on the Criteo dataset at a small batch size. When scaling to a large batch size in Table~\ref{tab:lb_criteo_128k} and Table~\ref{tab:lb_avazu_128k}, instead of AUC loss, our algorithm achieves a further performance gain of about $0.1\%$ in the Criteo. For Criteo-seq and Avazu, CowClip can scale up to 128$\times$ and 64$\times$ batch size without performance loss respectively. Equipped with CowClip, it can scale all four models to a large batch size with performance improvement, as shown in Table~\ref{tab:lb_criteo_128k} and Table~\ref{tab:lb_avazu_128k}. This shows that CowClip is a model-agnostic optmization technique.

Training with large batches can significantly reduce the training time. As shown in Table~\ref{tab:lb_criteo_time} for the Criteo dataset and Table~\ref{tab:lb_avazu_time} in Appendix for the Avazu dataset, the speedup achieved by scaling up the batch size is almost linear when the batch size is under 16K. We can still accomplish a sublinear speedup when continuing to scale up the batch size and achieve a 76.8$\times$ speed up with 128K batch size on the Criteo dataset. The compared four methods take advantage of different system optimization, such as reducing the communication and computational cost, so they achieve a much faster training speed with a 1K batch size. However, these methods have a low AUC performance and can only scale up to a 4K batch size due to performance loss in a larger batch size. Besides, they scale the batch size by using more GPUs, with 2 and 4 GPUs for 2K and 4K batch sizes, resulting in 2$\times$ and 4$\times$ cost in the GPU hours. In contrast, our method takes the advantage of large batch training, which can achieve a much shorter training time within only one GPU resource and obtain a higher AUC score.

\subsection{Ablation study}

Next, we show the superiority of CowClip over other clipping method designs with DeepFM on the Criteo dataset at batch size 8K and 128K. Table~\ref{tab:abl} gives the ablation of different gradient clipping designs. GC means the traditional gradient norm clipping. While it can help boost the performance when batch size equals 8K, it fails with b=128K. For the embedding table, we have two granularity: field and column (e.g., "Device" is a field, and "Mobile", "Computer" are columns). Field-wise GC and Column-wise GC show that gradient clipping on fine-grained granularity yields better results. The next two lines add the adaptive design to the clipping on the above two granularities, which adaptively decide the clipping values for each column (line 8 in Alg. 1). The reason that Field-wise adaptive GC fails to achieve a good result is because magnitudes of column gradients are different even in a field. Thus, Gradient clipping applied to a smaller unit yields better performance. CowClip (Adaptive Column-wise GC) outperforms all other methods in both settings. Hyperparameters for these clipping variants and more ablation study into the effectiveness of each component of CowClip can be found in the Appendix~\ref{app:more_abl}, which shows each component contributes to the final results.

\begin{table}[t]
\caption{Ablation study of CowClip on Criteo with DeepFM.}
\centering
\resizebox{\columnwidth}{!}{%
\begin{tabular}{@{}lcccc@{}}
\toprule
 & \multicolumn{2}{c}{\bf b = 8K} & \multicolumn{2}{c}{\bf b = 128K} \\ \cmidrule(lr){2-3}\cmidrule(lr){4-5}
 & AUC (\%) & LogLoss & AUC (\%) & LogLoss \\ \midrule
Gradient Clipping (GC) & 80.63 & 0.4452 & 77.24 & 0.4953 \\
Field-wise GC & 80.63 & 0.4453 & 80.62 & 0.4454 \\
Column-wise GC & 80.65 & 0.4095 & 80.75 & 0.4432 \\
Adaptive Field-wise GC & 80.62 & 0.4453 & 77.90 & 0.4824 \\ \midrule
Adaptive Column-wise GC & 80.97 & 0.4425 & 80.90 & 0.4430 \\ \bottomrule
\end{tabular}%
}
\label{tab:abl}
\end{table}

\section{Conclusion}

To accelerate the training of CTR prediction models on one GPU, we have explored large batch training and found that different frequencies hinder the scaling of learning rate and L2-regularization weight when scaling the batch size. Since previous scaling rules used in CV and NLP fail, we propose a novel optimization strategy CowClip with a simple scaling rule to stabilize the training process for large batch training in CTR prediction system. Experiments show that our method successfully scales the batch size to the state-of-the-art number and achieves significant training speedup. Our CowClip algorithm is also applicable to other tasks with a large embedding table such as NLP tasks.

\section*{Acknowledgement}

We thank Google TFRC for supporting us to get access to the Cloud TPUs. 
This work is supported NUS startup grant, the Singapore MOE Tier-1 grant, and the ByteDance grant.

\bibliography{aaai23}

\appendix
\onecolumn
\noindent\textbf{\Large Appendix}

\section{More Related Work}
\label{app:wdm}

\paragraph{Details on Wide/Cross stream of CTR prediction models}

Deep learning has been widely used by the community to boost CTR prediction performance. 
CTR prediction networks~\cite{Zhang2016DeepLO,Cheng2016WideD,Qu2016ProductBasedNN,Guo2018DeepFMAE,Wang2017DeepC,Lian2018xDeepFMCE,Song2019AutoIntAF,Li2019FiGNNMF,Deng2021DeepLightDL,Wang2021DCNVI,Wang2020CorrectNM,Wang2021MaskNetIF,Lyu2021MemorizeFO,Zhang2019FATDeepFFMFA,Zhang2018DeepLB} have outperformed traditional methods such as Logistic Regression (LR)~\cite{McMahan2013AdCP} and Factorization Machine (FM)~\cite{Rendle2010FactorizationM}. 
The progress of AUC of CTR prediction models on the Criteo dataset is shown in Figure~\ref{fig:f1_left}.

Here we illustrate the architecture of models used in our experiment. After one-hot encoding of every categorical field, the input is embedded into a dense vector in the network. The wide or cross-stream serves to model the feature interactions explicitly. For instance, the LR model is a first-order predictor, FM models the second-order interactions, and $n$ cross layers~\cite{Wang2017DeepC} models $n$-th-order interactions. The first two methods are called ``wide'' for only one layer is used. The other stream is a feed-forward neural network to compensate for the ability to learn higher-order interactions.

The DeepFM~\cite{Guo2018DeepFMAE} adopts a factorization machine for the wide part, which is also the cross-stream for it explicitly models the second-order relationship between different features. After embedding each column into a $d$-dimensional vector $\bm{v}$, the factorization machine models $\hat y$ as follows:
\begin{equation*}
    \hat y=w_0+\sum_{i=1}^nw_i\bm{x}_i+\sum_{i=1}^n\sum_{j=i+1}^n\innerproduct{\bm{v}_i}{\bm{v}_j}\bm{x}_i\bm{x}_j.
\end{equation*}

For the W\&D model, the wide part is logistic regression. Denote the output of the network as $\hat y$, the predicted probability of clicking is obtained by $\texttt{sigmoid}(\hat y)$. After one-hot encoding of input $x$, the vector is a $n$-dimensional $\bm{x}$. The logistic regression considers each feature independently and lets the DNN captures high-order relations in the data.
\begin{equation*}
    \hat y=w_0+\sum_{i=1}^nw_i\bm{x}_i.
\end{equation*}
For DCN~\cite{Wang2017DeepC}, it introduces the cross-layer to automatically learns a high-order interaction. With $L$ cross layers, it can model interactions from $2$ to $L+1$ order. Denote the $\bm{x}_0$ as input vector and $\bm{x}_\ell$ as the output of the $\ell$-th layer. The $\ell$-th cross-layer does as follows:
\begin{equation*}
    \bm{x}_{\ell+1}=\bm{x}_0\bm{x}_{\ell}^\top\bm{w}_{\ell}+\bm{b}_{\ell}+\bm{x}_{\ell}.
\end{equation*}
The DCN-v2~\cite{Wang2021DCNVI} proposed a new cross-layer to model high-order interaction. Specifically, with $\bm{W}_{\ell}\in\mathbb{R}^{d\times d}$, the $\ell$-th layer is:
\begin{equation*}
    \bm{x}_{\ell+1}=\bm{x}_0\odot(\bm{W}_{\ell}\bm{x}_{\ell}+\bm{b}_{\ell})+\bm{x}_{\ell}.
\end{equation*}

\begin{figure}[t]
    \centering
    \includegraphics[width=.5\linewidth]{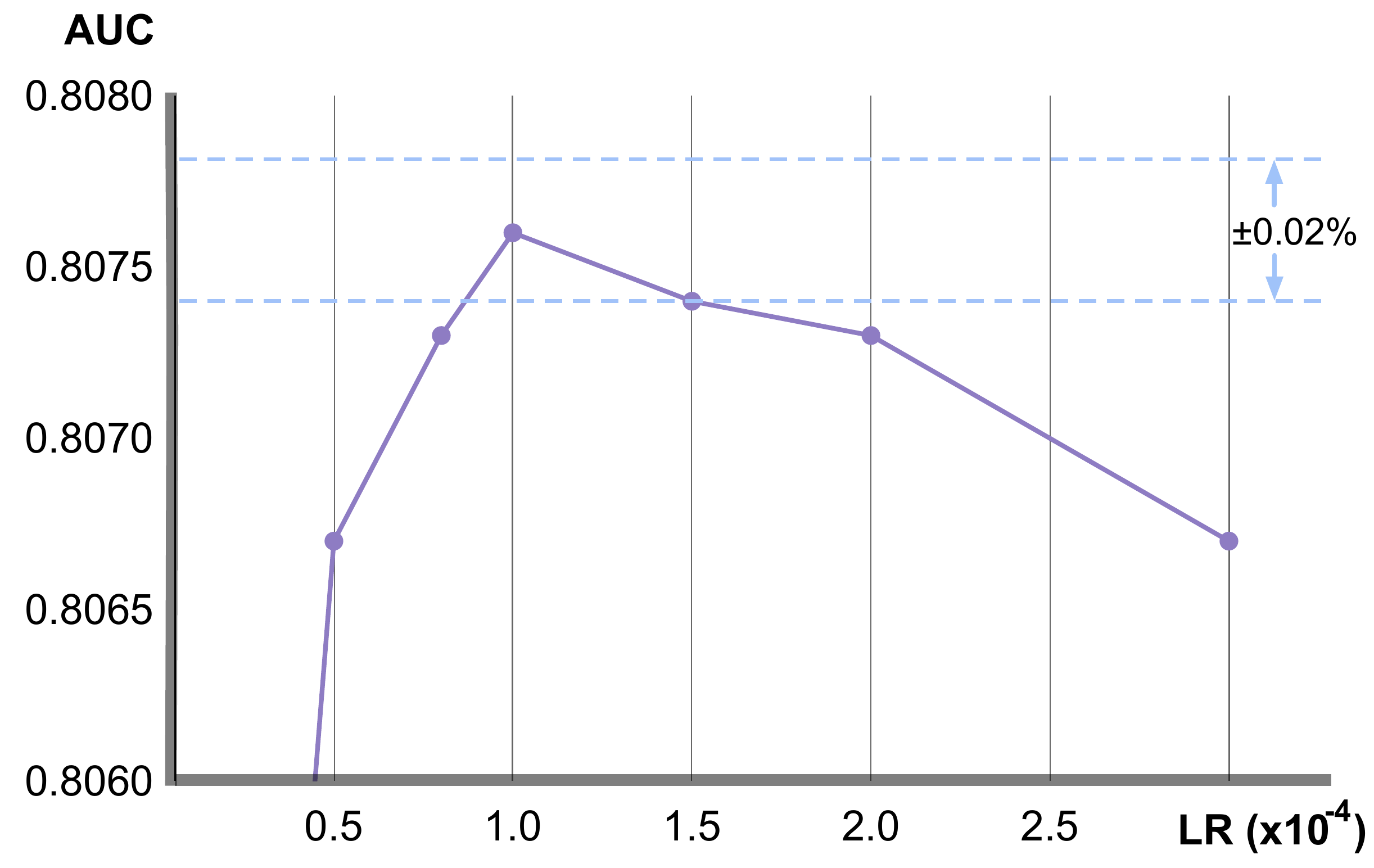}
    \caption{AUC performance of DeepFM on Criteo dataset with different learning rates. The one-year improvement is about 0.3\%. The accuracy drops a lot when a slight disturbance is made.}
    \label{fig:f1_right}
\end{figure}

\paragraph{Sensitiveness of CTR prediction.}
Preserving the CTR prediction model's performance with a large batch size is a great challenge in that the CTR model is very sensitive. In accelerating the training of ResNet-50, an accuracy loss within 1\% is tolerable~\cite{You2017LargeBT,You2020LargeBO, Kumar2019ScaleMM}. However, considering the tremendous amounts of clicking happening every day, a 0.1\% loss in AUC will cost a company too much to bear. As shown in Figure~\ref{fig:f1_left}, a continuous effort to develop new CTR prediction models only improved the AUC by less than 2\% on the Criteo~\cite{criteo} dataset in the past six years., and an improvement in a month of 0.02\% is considered significant in Criteo dataset in this paper.
In our experiments, a tiny shift in learning rate results in a significant 0.02\% drop in model performance at batch size 1K (Figure~\ref{fig:f1_right}), indicating the task is sensitive to hyperparameters.

\paragraph{Methods utilizing embedding frequency in CTR prediction}
\label{app:freq}
Many works have shed light on the importance of different frequencies in embeddings to accelerate or improve the accuracy of CTR prediction models.
\cite{Miao2021HETSO,Adnan2021AcceleratingRS} finds that most access to the embedding layer happens in a small fraction of embeddings. They both propose to cache the most frequent embeddings to reduce the communication bottleneck.~\cite{Ginart2021MixedDE} proposed a mixed dimension embedding where the embedding vector’s dimension scales with its query frequency to reduce the memory cost.
To improve the performance of the CTR prediction models,~\cite{Zhou2019DeepIE} shows a regularization by filtering out ids with a low occurrence frequency.~\cite{Li2016DiFactoDF} proposes to apply stronger regularization on more frequent ids for better generalization performance. In this paper, we show that the different frequencies of ids result in the failure of previous scaling rules when scaling up the batch size.

\section{Traditional Scaling Rules Derivation}
\label{app:tradderi}

The deduction of scaling rules is mostly based on the SGD optimizer. However, many experimental results~\cite{He2021LargeScaleDL,You2020LargeBO,Hoffer2017TrainLG} show they are also effective when applied to adaptive optimizers such as Adam. We consider the case batch size is scaled from $B$ to $B'$, where $b'=|B'|=s|B|=sb$. For a given big batch $B'_t$, the corresponding $s$ small batches are $\{B_{t,i}\}_{i=1}^s$.

\paragraph{Linear Scaling.}
For Linear Scaling, when scaling up $s$ times the batch size, the motivation is to keep the update by a big batch equivalent to the update made within $s$ small batches. Denote the update made by SGD update as $\Delta w = w_t-w_{t+1}$, for the small batches. We have the expected update:
\begin{equation*}
    \\E[\Delta w] = \eta\sum_{i=1}^{s}\E[\frac{1}{b}\cdot\sum_{x\in B_{t,i}}\nabla L(w_{t,i}, x)]=\eta\sum_{i=1}^s \E[\nabla L(w_{t,i}, x)],
\end{equation*}
while for the big batch:
\begin{equation}
\E[\Delta w] = \eta' \E[\frac{1}{b'}\cdot\sum_{x\in B_t'}\nabla L(w_t, x)]=\eta'\E[\nabla L(w_t, x)].
\end{equation}
Under the assumption that $\E[\nabla L(w_{t,i}, x)]\approx \E[\nabla L(w_{t}, x)]$ for $i=1$ to $s$, to make the two update equal, we need to scale the learning rate $\eta' \rightarrow s\eta$.

\paragraph{Sqrt Scaling.}
For Sqrt Scaling, its motivation is to keep the covariance matrix of the parameters updates $\Delta w$ the same. The derivation comes from~,\cite{Hoffer2017TrainLG} and we only consider the case that samples are randomly drawn from a dataset with replacement. According to~\cite{Hoffer2017TrainLG}, with minibatch gradient denoted as $\bm{\hat g}=\frac{1}{b} \sum_{x\in B}\bm{g}_x$, we have
\begin{equation*}
    \texttt{cov}(\bm{\hat g}, \bm{\hat g}) = (\frac{1}{b}-\frac{1}{N})\frac{1}{N}\sum_{i=1}^N \bm{\hat g}_i\bm{\hat g}_i^\top.
\end{equation*}
Since $\frac{1}{N}$ is small, the update of SGD has covariance:
\begin{equation}
    \label{eq:sqrtcov}
    \texttt{cov}(\Delta w, \Delta w) = \texttt{cov}(\eta\bm{\hat g}, \eta\bm{\hat g}) \approx \frac{\eta^2}{b\cdot N}\sum_{i=1}^N \bm{\hat g}_i\bm{\hat g}_i^\top.
\end{equation}
When scaling $b\rightarrow sb$, to keep the covariance matrix to be the same, we need to scale $\eta\rightarrow \sqrt{s}\eta$.

\paragraph{Scaling the L2-regularization weight.}

We follow the discussion in~\cite{Krizhevsky2014OneWT} to scale the L2-regularization weight with batch size. Consider the case only L2-regularization is applied, under a large batch we have:
\begin{equation*}
    w_{t+1} = w_t(1-\eta'\lambda' w),
\end{equation*}
while for small batches:
\begin{equation*}
    w_{t+1} = w_t(1-\eta\lambda w)^s= w_t(1-s\eta\lambda w).
\end{equation*}
To make the L2-regularization strength at the same level, when scaling the batch size, we have the equation:
\begin{equation*}
    \eta'\lambda'\approx s\eta\lambda.
\end{equation*}
Therefore, considering the learning rate scaling in linear scaling and sqrt scaling, it is easy to calculate the scaling strategy for the $\lambda$.

\section{Sqrt Scaling Motivation with Different Frequencies}
\label{app:sqrt}

In Section~\ref{subsec:fail}, we have discussed the scaling rules for columns with different frequencies. Here, we consider the sqrt scaling rule. The derivation is inspired by~\cite{Hoffer2017TrainLG}. Similar to Equation~\ref{eq:pe}, we have
\begin{equation*}
    \bm{g} = \frac{1}{b}\delta(\text{id}_k^{f_j}\in B)\sum_{x\in B}\nabla L(w, x)=\delta(\text{id}_k^{f_j}\in B)\bm{\hat g}.
\end{equation*}
With Equation~\ref{eq:sqrtcov}, since mini-batches are uncorrelated, the covariance is
\begin{align*}
    \texttt{cov}(\Delta \bm{g}, \Delta \bm{g}) & 
    = \E[\bm{g}\bm{g}^\top]-\E[\bm{g}]\E[\bm{g}^\top]\\
    & = \frac{1}{b^2}\sum_{i=1}^b\sum_{j=1}^b \E[\delta(\text{id}_k^{f_j}\in B)\delta(\text{id}_k^{f_j}\in B')] \bm{\hat g}_i\bm{\hat g}_j - \prob(\text{id}_k^{f_j}\in B)^2\E[\bm{\hat g}]\E[\bm{\hat g^\top}] \\
    & = \prob(\text{id}_k^{f_j}\in B) \E[\bm{\hat g}\bm{\hat g^\top}] - \prob(\text{id}_k^{f_j}\in B)^2\E[\bm{\hat g}]\E[\bm{\hat g^\top}].
\end{align*}
As we can see, the frequency differently scales the two parts in covariance. It is impossible to correct to the original behavior by multiplying a scaler  to the learning rate. Only under a strong assumption that $\E[\bm{\hat g}\bm{\hat g^\top}]\ll \E[\bm{\hat g}]\E[\bm{\hat g^\top}]$ can we make the following approximation:
\begin{equation*}
     \texttt{cov}(\Delta \bm{g}, \Delta \bm{g}) \approx \prob(\text{id}_k^{\text{f}_j}\in B)\texttt{cov}(\Delta \bm{\hat g}, \Delta \bm{\hat g}).
\end{equation*}
The assumption may not hold in practice, which adds to the problem's difficulty. Under the hypothesis, we have
the covariance matrix for the $\Delta w$ is:
\begin{equation*}
    \texttt{cov}(\Delta w, \Delta w) = \texttt{cov}(\eta\bm{\hat g}, \eta\bm{\hat g}) \approx \prob(\text{id}_k^{\text{f}_j}\in B)\cdot\frac{\eta^2}{b\cdot N}\sum_{i=1}^N \bm{\hat g}_i\bm{\hat g}_i^\top.
\end{equation*}
For dense weight and frequent ids, with $\prob(\text{id}_k^{\text{f}_j}\in B)\approx 1$, it is the same to the sqrt scaling. However, for infrequent ids, the covariance matrix becomes:
\begin{equation*}
    \texttt{cov}(\Delta w, \Delta w) = \texttt{cov}(\eta\bm{\hat g}, \eta\bm{\hat g}) \approx \frac{\eta^2}{\freq\cdot N}\sum_{i=1}^N \bm{\hat g}_i\bm{\hat g}_i^\top.
\end{equation*}
Hence, there is no need to scale the learning rate to keep the same covariance matrix.

% \begin{align*}
%     \texttt{cov}(\bm{\hat g}, \bm{\hat g}) & = \E[\bm{\hat g}\bm{\hat g}^\top] - \E[\bm{\hat g}]E^\top[\bm{\hat g}] \\
% \end{align*}

% \section{Convergence of CowClip Algorithm}
% \label{app:conv}

\section{Effect of Loss Scaling}
\label{app:adam}

In training the network, the gradient of weights $w$ on one sample $x$ concerning the training loss $L$ is $\nabla_{w} L(w, x)$. When training with batch size $b$, we take the average of gradients on the samples and thus have the expected gradients as follows:
\begin{equation*}
    \E[G]=\E[\frac{1}{b}\sum_{i=1}^b\nabla_{w} L(w, x_i)]=\E[\nabla_{w} L(w, x)].
\end{equation*}
As discussed in Equation~\ref{eq:pe}, in training with weights of different frequencies, there is a case a constant multiplier is applied to the expected gradients. We consider the case gradients are scaled by $c$ and show different behaviors with SGD and Adam with L2-regularization following discussion in~\cite{stack395443}:
\begin{equation*}
    \E[G] = c\cdot \E[\nabla_{w} L(w, x)].
\end{equation*}
First, for SGD optimizer with learning rate $\eta$ and L2-regularization weight $\lambda$, the expected update is:
\begin{equation*}
    \E[\Delta w] = \E[w_t-w_{t+1}] = \eta\cdot(\E[\nabla_{w_t} L(w_t, x)] - \frac{\lambda}{2}w_t).
\end{equation*}
When the gradients are scaled by $c$, we have
\begin{equation*}
    \E[\Delta w] = \eta\cdot(c\cdot \E[\nabla_{w_t} L(w_t, x)] - \frac{\lambda}{2}w_t) = c\eta\cdot(\cdot \E[\nabla_{w_t} L(w_t, x)] - \frac{\lambda}{2c}w_t).
\end{equation*}
So the effect is using a new learning rate of $c\eta$ and new L2-regularization weight of $\frac{\lambda}{c}$.

For Adam optimizer with $(\beta_1,\beta_2)$ without L2-regularization, the expected update is:
\begin{equation}
    \E[\Delta w] = \eta \E[\frac{m_t}{\sqrt{v_t+\epsilon}}].
\end{equation}
If we omit the $\epsilon$ term and bias correction for simplicity, when the gradients are scaled by $c$, the momentum term $m_t$ has:
\begin{equation*}
    \E[m_t] = \beta_1 \E[m_t] + (1-\beta_1)\cdot c\cdot \E[\nabla_{w_t} L(w_t, x)] = c \cdot \sum_{i=1}^t (1-\beta_1)\beta_1^{i-1} \E[\nabla_{w_i} L(w_i, x)].
\end{equation*}
which is $c$ times the original $\E[m_t]$. The similar deduction finds $v_t\rightarrow c^2 v_t$. Thus,
\begin{equation*}
    \E[\Delta w]=\frac{c\cdot m_t}{\sqrt{c^2\cdot v_t+\epsilon}}\approx\frac{m_t}{\sqrt{v_t+\epsilon}},
\end{equation*}
so the behaviour of Adam without regularization is not changed. However, with L2-regularization, we have:
\begin{equation*}
\E[m_t] = c \cdot \sum_{i=1}^t (1-\beta_1)\beta_1^{i-1} (\E[\nabla_{w_i} L(w_i, x)]+\frac{\lambda}{2c} w_t),
\end{equation*}
and the same deduction applies to $v^2$. This shows that when the gradients are scaled by $c$, it is equivalent to using a new L2-regularization weight of $\frac{\lambda}{c}$.

% \usepackage{booktabs}
% \usepackage{graphicx}
\begin{table}[t]
\centering
\caption{hyperparameters for square root scaling, linear scaling and empirical scaling.}
% \resizebox{\textwidth}{!}{%
\begin{tabular}{@{}lccccccc@{}}
\toprule
 & \multicolumn{2}{c}{\textbf{Sqrt Scaling}} & \multicolumn{2}{c}{\textbf{Linear Scaling}} & \multicolumn{3}{c}{\textbf{Empirical Scaling}} \\\cmidrule(lr){2-3}\cmidrule(lr){4-5}\cmidrule(lr){6-8}
{\bf Batch Size} & LR & L2 & LR & L2 & LR (Embed) & L2 & LR (Dense) \\ \midrule
1K (1024) & $1\times 10^{-4}$ & $1\times 10^{-4}$ & $1\times 10^{-4}$ & $1\times 10^{-4}$ & $1\times 10^{-4}$ & $1\times 10^{-4}$ & $1\times 10^{-4}$ \\
2K (2048) & $\sqrt{2}\times 10^{-4}$ & $\sqrt{2}\times 10^{-4}$ & $2\times 10^{-4}$ & $1\times 10^{-4}$ & $1\times 10^{-4}$ & $4\times 10^{-4}$ & $2\times 10^{-4}$ \\
4K (4096) & $2\times 10^{-4}$ & $2\times 10^{-4}$ & $4\times 10^{-4}$ & $1\times 10^{-4}$ & $1\times 10^{-4}$ & $1.6\times 10^{-3}$ & $4\times 10^{-4}$ \\
8K (8192) & $2\sqrt{2}\times 10^{-4}$ & $2\sqrt{2}\times 10^{-4}$ & $8\times 10^{-4}$ & $1\times 10^{-4}$ & $1\times 10^{-4}$ & $\underline{1.28\times 10^{-2}}$ & $8\times 10^{-4}$ \\ \bottomrule
\end{tabular}%
% }
\label{tab:hp1}
\end{table}

% \usepackage{booktabs}
% \usepackage{graphicx}
\begin{table}[t]
\centering
\caption{hyperparameters for CowClip scaling on Criteo and Avazu dataset.}
\resizebox{\textwidth}{!}{%
\begin{tabular}{@{}lcccccccc@{}}
\toprule
 & \multicolumn{4}{c}{\textbf{Criteo}} & \multicolumn{4}{c}{\textbf{Avazu}} \\\cmidrule(lr){2-5}\cmidrule(lr){6-9}
{\bf Batch Size} & LR (Embed) & L2 & LR (Dense) & $(r, \zeta)$ & LR (Embed) & L2 & LR (Dense) & $(r, \zeta)$ \\ \midrule
1K (1024) & $1\times 10^{-4}$ & $1\times 10^{-4}$ & $8\times 10^{-4}$ & $(1, 10^{-5})$ & $1\times 10^{-4}$ & $1\times 10^{-4}$ & $1\times 10^{-4}$ & $(10, 10^{-3})$ \\
2K (2048) & $1\times 10^{-4}$ & $2\times 10^{-4}$ & $8\sqrt{2}\times 10^{-2}$ & $(1, 10^{-5})$ & $1\times 10^{-4}$ & $2\times 10^{-4}$ & $\sqrt{2}\times 10^{-4}$ & $(10, 10^{-3})$ \\
4K (4096) & $1\times 10^{-4}$ & $4\times 10^{-3}$ & $16\times 10^{-4}$ & $(1, 10^{-5})$ & $1\times 10^{-4}$ & $4\times 10^{-3}$ & $2\times 10^{-4}$ & $(1, 10^{-4})$ \\
8K (8192) & $1\times 10^{-4}$ & $8\times 10^{-4}$ & $16\sqrt{2}\times 10^{-4}$ & $(1, 10^{-5})$ & $1\times 10^{-4}$ & $8\times 10^{-4}$ & $2\sqrt{2}\times 10^{-4}$ & $(1, 10^{-4})$ \\
16K (16384) & $1\times 10^{-4}$ & $1.6\times 10^{-3}$ & $32\times 10^{-4}$ & $(1, 10^{-5})$ & $1\times 10^{-4}$ & $1.6\times 10^{-3}$ & $4\times 10^{-4}$ & $(1, 10^{-4})$ \\
32K (32768) & $1\times 10^{-4}$ & $3.2\times 10^{-3}$ & $32\sqrt{2}\times 10^{-4}$ & $(1, 10^{-5})$ & $1\times 10^{-4}$ & $3.2\times 10^{-3}$ & $4\sqrt{2}\times 10^{-4}$ & $(1, 10^{-4})$ \\
64K (65536) & $1\times 10^{-4}$ & $6.4\times 10^{-3}$ & $64\times 10^{-4}$ & $(1, 10^{-5})$ & $1\times 10^{-4}$ & $6.4\times 10^{-3}$ & $8\times 10^{-4}$ & $(1, 10^{-4})$ \\
128K (131072) & $1\times 10^{-4}$ & $1.28\times 10^{-2}$ & $64\sqrt{2}\times 10^{-4}$ & $(1, 10^{-5})$ & $1\times 10^{-4}$ & $\underline{9.6\times 10^{-3}}$ & $\underline{16\times 10^{-4}}$ & $(1, 10^{-4})$ \\ \bottomrule
\end{tabular}%
}
\label{tab:hp2}
\end{table}

\section{Additional Implementation Details}
\label{app:impl}

In this paper, ``K'' means $\times 1024$, so 1K means 1024 batch size. Following the common network setting~\cite{Guo2018DeepFMAE}, the dimension of categorical field embedding is 10, the depth of hidden layers for MLP is 3, and the number of neurons is 400 per layer. All models are trained with 10 epochs, and the final model is evaluated at the test set. The detailed learning rate, L2-regularization weights, and other hyperparameters are listed in Table~\ref{tab:hp1} and Table~\ref{tab:hp2}, square roots are round to four decimal places in practice. The hyperparameter $r$ is not sensitive, so we set it directly to $1$, and the choice of $\zeta$ is related to the initialization weight in the next paragraph. LR denotes the learning rate $\eta$, and L2 denotes the L2-regularization weight $\lambda$ (no L2-regularization is imposed on dense weights). All activation functions are ReLU~\cite{agarap2018deep}, and dropout~\cite{Srivastava2014DropoutAS} is not used as we do not see its improvement. For DCN and DCNv2, the number of cross-layer is 3, and we only adopt the cross-layer form from the DCNv2 paper. The CowClip method is performed on id vector embeddings (columns), but is not applied to LR method for DeepFM and W\&D, whose biases can be viewed as a 1-dimension embedding. For the continuous field, they do not involve in the wide or cross stream and are directly sent into the DNN stream.

One technique to train with CowClip is that as the learning process of embedding becomes more smooth and stable, we can fix the learning rate for the embeddings and scale up the learning rate for the dense layer until the training process diverges for better performance. As the batch size grows beyond a threshold, the proposed scaling rule may face accuracy loss (8K for empirical scaling rules and 128K for the CowClip scaling rule in the Avazu dataset). In that case, we do a little hyperparameter fine-tuning by scaling some of the hyperparameters to twice or half of their supposed value. Basically, we increase the L2 $\lambda$ when the network is overfitting and increase $\zeta$ when the network is underfitting. These values are underlined in the table.

% fig: training loss & auc curve
\begin{figure}[t]
     \centering
     \includegraphics[width=\textwidth,trim={0pt 3.8cm 0pt 6cm},clip]{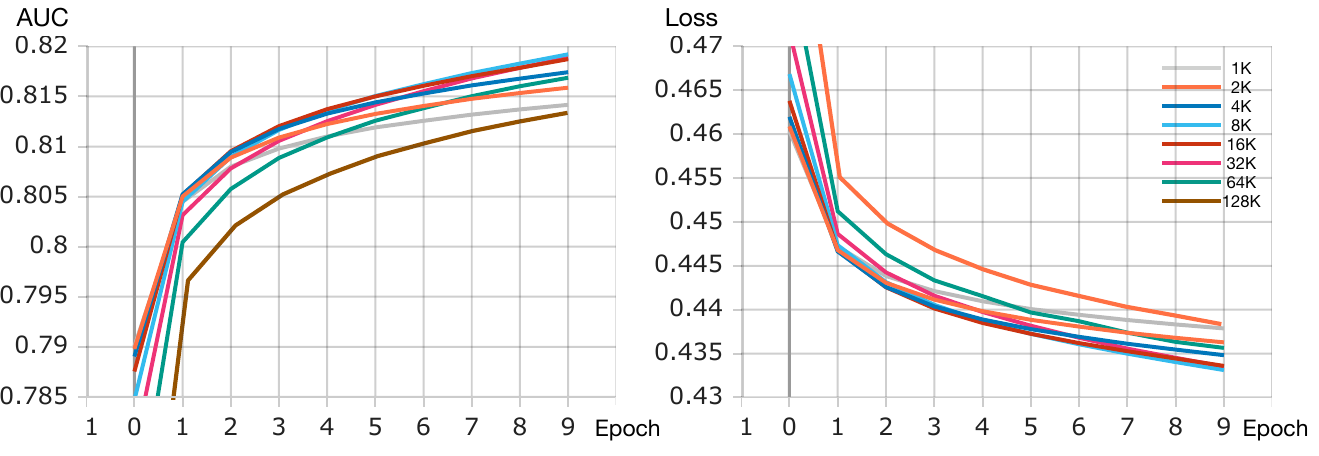}
    \caption{Training AUC (left) and Loss (right) at different epochs with different batch sizes during the training.}
    \label{fig:train}
\end{figure}
\begin{figure}[!t]
     \centering
     \includegraphics[width=\textwidth]{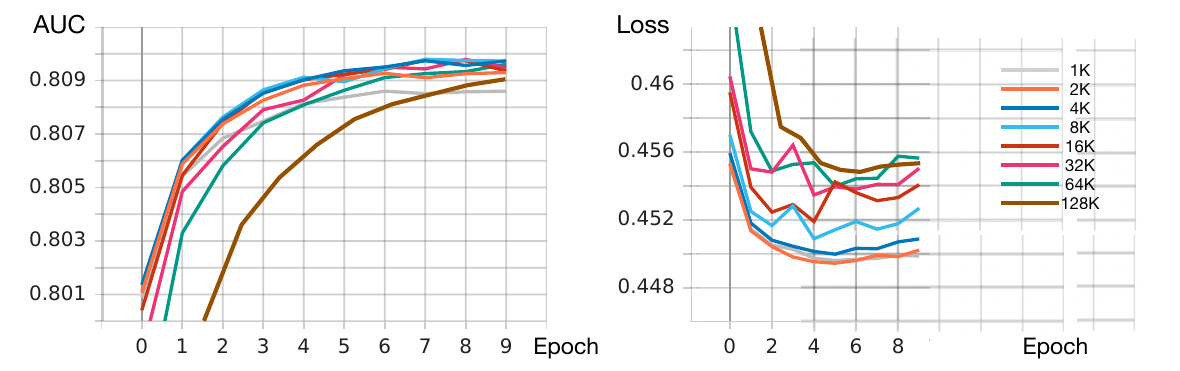}
    \caption{Test AUC (left) and Loss (right) at different epochs with different batch sizes during the training.}
    \label{fig:test}
\end{figure}

For large batch training with CowClip, there are two additional techniques. Warmup on learning rate has been widely used in training CV and NLP networks~\cite{Gotmare2019ACL}. It helps the network start smoothly for a more stable training process. We find warmup on the learning rate of the embeddings has little improvement, so we only apply one-epoch learning rate warmup to the dense weights of the CTR prediction model. Weight initialization is also important for a good starting state. We use Kaiming initialization~\cite{He2015DelvingDI} for all dense weights. The original initialization for embeddings is $w\sim \mathcal{N}(0, \sigma),\sigma=10^{-4}$. With a dimension of $d$, the initial weight norm is $\sqrt{d}\cdot\sigma$. To allow for a greater gradient norm bound in CowClip, we use a larger initial weight by setting the $\sigma$ to $10^{-2}$ for training with CowClip. To avoid a too strong clipping value, we lower bound the clipping value with $10^{-4}$, which is the original initial weight norm, and in Criteo $10^{-5}$, which yields a better result.

\section{Additional Experimental Results}

We run our experiments with three random seeds (1234, 1235, 1236), and the standard deviation among all experiments for AUC is less than 0.012\%. The performance comparisons between previous scaling methods and CowClip for Criteo-seq and Avazu dataset are shown in Table~\ref{tab:lb_criteo_seq_8k} and Table~\ref{tab:lb_avazu_8k} respectively. The performance of CowClip methods at different batch size with four different network architectures on Avazu is presented in Table~\ref{tab:lb_avazu_128k}. As we can see, Cowclip maintains the performance at a large batch size on both dataset and achieves a fast speedup (for Criteo-seq, the speedup is the same as Criteo). The training time comparison and speedup for the Avazu dataset are shown in Table~\ref{tab:lb_avazu_time}.The training and testing AUC and loss curve at different epochs with different batch sizes are shown in Figure~\ref{fig:train} and Figure~\ref{fig:test} respectively.

\begin{table}[!t]
\caption{AUC (\%) of different scaling methods on Criteo-seq dataset from 1K to 8K on DeepFM.}
\centering
% \resizebox{0.8\textwidth}{!}{%
\begin{tabular}{@{}lccccc@{}}
\toprule
 & {\bf 1K} & {\bf 2K} & {\bf 4K} & {\bf 8K} & {\bf 128K} \\\midrule
No Scaling & 80.48 & 80.25 & 79.79 & 79.04 & --\\
Sqrt Scaling & 80.48 & 80.26 & 79.82 & 79.91 & --\\
LR Scaling & 80.48 & 80.29 & 80.29 & 80.03 & --\\
CowClip (Ours) & 80.48 & 80.50 & 80.50 & 80.49 & 80.49\\ \bottomrule
\end{tabular}%
% }
\label{tab:lb_criteo_seq_8k}
% \vspace{-5pt}
\end{table}

\begin{table}[!t]
\caption{Performance of different scaling methods on Avazu dataset from 1K to 8K on DeepFM.}
\centering
\resizebox{\textwidth}{!}{%
\begin{tabular}{@{}lcccccccc@{}}
\toprule
 & \multicolumn{2}{c}{\bf 1K (1024)} & \multicolumn{2}{c}{\bf 2K (2048)} & \multicolumn{2}{c}{\bf 4K (4096)} & \multicolumn{2}{c}{\bf 8K (8192)} \\\cmidrule(lr){2-3}\cmidrule(lr){4-5}\cmidrule(lr){6-7}\cmidrule(lr){8-9}
 & AUC (\%) & LogLoss & AUC (\%) & LogLoss & AUC (\%) & LogLoss & AUC (\%) & LogLoss \\ \midrule
No Scaling & 78.84 & 0.3748 & 78.79 & 0.3775 & 77.69 & 0.3952 & 75.85 & 0.4411 \\
Sqrt Scaling & 78.84 & 0.3748 & 78.88 & 0.3761 & 77.78 & 0.3926 & 76.23 & 0.4299 \\
Sqrt Scaling$^*$ & 78.84 & 0.3748 & 78.88 & 0.3759 & 77.98 & 0.3976 & 76.23 & 0.4140 \\
LR Scaling & 78.84 & 0.3748 & 78.78 & 0.3763 & 77.72 & 0.3883 & 76.69 & 0.4043 \\
$n^2$--$\lambda$ Scaling (Ours) & 78.84 & 0.3748 & 78.84 & 0.3754 & 78.26 & 0.3815 & 77.24 & 0.3912 \\
CowClip (Ours) & 78.83 & 0.3748 & 78.82 & 0.3752 & 78.90 & 0.3752 & 79.06 & 0.3740 \\ \bottomrule
\end{tabular}%
}
\label{tab:lb_avazu_8k}
\end{table}

\begin{table}[!t]
\caption{Performance of CowClip methods on Avazu dataset from 1K to 128K on four models.}
\centering
\resizebox{\textwidth}{!}{%
\begin{tabular}{@{}clccccccccc@{}}
\toprule
 &  & Baseline & 1K & 2K & 4K & 8K & 16K & 32K & 64K & 128K \\ \midrule
\multirow{2}{*}{DeepFM~\cite{Guo2018DeepFMAE}} & AUC (\%) & 78.84 & 78.83 & 78.82 & 78.90 & 79.06 & 79.01 & 78.82 & 78.82 & 78.80 \\
 & Logloss & 0.3748 & 0.3751 & 0.3752 & 0.3752 & 0.3740 & 0.3759 & 0.3780 & 0.3781 & 0.3758 \\ \midrule
\multirow{2}{*}{W\&D~\cite{Cheng2016WideD}} & AUC (\%) & 78.80 & 78.80 & 78.81 & 78.90 & 79.06 & 79.03 & 78.82 & 78.81 & 78.79 \\
 & Logloss & 0.3752 & 0.3754 & 0.3752 & 0.3752 & 0.3744 & 0.3754 & 0.3782 & 0.3784 & 0.3758 \\ \midrule
\multirow{2}{*}{DCN~\cite{Wang2017DeepC}} & AUC (\%) & 78.82 & 78.80 & 78.81 & 78.91 & 79.05 & 78.97 & 78.74 & 78.78 & 78.79 \\
 & Logloss & 0.3749 & 0.3754 & 0.3752 & 0.3751 & 0.3744 & 0.3760 & 0.3787 & 0.3780 & 0.3758 \\ \midrule
\multirow{2}{*}{DCN v2~\cite{Wang2021DCNVI}} & AUC (\%) & 78.84 & 78.83 & 78.82 & 78.89 & 79.07 & 78.97 & 78.80 & 78.81 & 78.75  \\
 & Logloss & 0.3748 & 0.3750 & 0.3754 & 0.3751 & 0.3742 & 0.3760 & 0.3778 & 0.3779 & 0.3760 \\ \bottomrule
\end{tabular}%
}
\label{tab:lb_avazu_128k}
\end{table}

\begin{table}[!t]
\caption{The training time of different methods on Avazu dataset. . Last four are trained with CowClip.}
\centering
\resizebox{\textwidth}{!}{%
\begin{threeparttable}
\begin{tabular}{@{}lcccccccccc@{}}
\toprule
 &  &  & \multicolumn{8}{c}{\bf Time (minutes)} \\\cmidrule(lr){4-11}
 & AUC (\%) & Logloss & 1K & 2K & 4K & 8K & 16K & 32K & 64K & 128K \\ \midrule
XDL~\cite{Adnan2021AcceleratingRS} & 75.8 & 0.390 & 108 & 84 & 74 & -- & -- & -- & -- & -- \\
FAE~\cite{Adnan2021AcceleratingRS} & 77.8 & 0.391 & 72 & 62 & 61 & -- & -- & -- & -- & -- \\
DLRM~\cite{Naumov2019DeepLR} & 76.6 & 0.387 & 163 & 141 & 54 & -- & -- & -- & -- & -- \\
Hotline~\cite{Adnan2021AcceleratingID} & 76.8 & 0.386 & 70 & 28 & 24 & -- & -- & -- & -- & -- \\ \midrule
DeepFM & 78.84 & 0.3748 & 210 & 108 & 54 & 30 & 17 & 10 & 6.7 & 4.8 \\
W\&D & 78.80 & 0.3750 & 210 & 108 &   54 & 30 & 17 & 10 & 6.7 & 5.0 \\
DCN & 78.82 & 0.3749 & 210 & 108  &   54 & 30 & 18 & 11 & 7.2 & 5.7 \\
DCN v2 & 78.84 & 0.3748 & 234 & 126  &   66 & 37 & 25 & 19 & 18.5 & 19.5 \\ \midrule
\multicolumn{2}{l}{\hspace{-6pt}Speedup (DeepFM)} &  & 1$\times$ & 1.94$\times$ & 3.89$\times$ & 7.00$\times$ & 12.3$\times$ & 21$\times$ & 31.3$\times$ & 43.7$\times$ \\ \bottomrule
\end{tabular}%
\begin{tablenotes}
    \footnotesize
    \item[$\dagger$] Trained with 2 GPUs $^\ddagger$ Trained with 4 GPUs.
\end{tablenotes}
\end{threeparttable}
}
\label{tab:lb_avazu_time}
\end{table}

% \usepackage{booktabs}
% \usepackage{graphicx}
% gc [25, 20, 10, 1] no difference
% column-wise gc [1e-5, 1e-4 (128K), 1e-3 (8K)]
% field-wise gc [0.001, 0.1 (128), 1, 10]
% field-wise agc [0.1, 1 (128), 10]
\begin{table}[!t]
\caption{More Ablation study of CowClip on Criteo with DeepFM.}
\centering
% \resizebox{\textwidth}{!}{%
\begin{tabular}{@{}lcccc@{}}
\toprule
 & \multicolumn{2}{c}{\bf b = 8K} & \multicolumn{2}{c}{\bf b = 128K} \\\cmidrule(lr){2-3}\cmidrule(lr){4-5}
 & AUC (\%) & LogLoss & AUC (\%) & LogLoss \\ \midrule
 CowClip w./ Linear Scale on Dense & diverge & diverge & diverge & diverge \\
CowClip w./ Empirical Scale & 80.85 & 0.4430 & 79.83 & 0.4539 \\
CowClip w.o. $\zeta$ & 80.96 & 0.4426 & 80.88 & 0.4438 \\
CowClip w.o. warmup & 80.97 & 0.4422 & 80.52 & 0.4463 \\
CowClip w.o. large init weight & 80.92 & 0.4432 & 80.90 & 0.4431 \\ \midrule
CowClip & 80.97 & 0.4425 & 80.90 & 0.4430 \\ \bottomrule
\end{tabular}%
% }
\label{tab:more_abl}
\end{table}

\section{More Ablation Study}
\label{app:more_abl}

More ablation study is shown in Table~\ref{tab:more_abl}. The first two rows verify the effectiveness of our scaling rule. Warmup on the dense weights is critical when the batch size is very large, while large initialization weights prevent the network from underfitting when the batch size is not that large.

We decide the hyperparameters for variants of gradient clipping in Table~\ref{tab:abl} as follows, which may be helpful if a simple version of gradient clipping is adopted for a complex system. First, we run the experiment and log out the gradients of interested units (i.e., global, field, column). For gradient clipping, the upper bound is $25$, and after searching in $\{25, 20, 10, 1\}$, we find the performance is not sensitive to the clipping value. For field-wise and column-wise, we search in $\{10^{-3}, 10^{-2}, 10^{-1}, 1, 10\}$ and $\{10^{-5}, 10^{-4}, 10^{-3}\}$ respectively.

For the gradient clipping with constant value $\texttt{clip\_t}$ for the global embedding or a field, note that when scaling the batch size, we also need to scale this value. Take the field-wise gradient clipping for example. Consider scaling the batch size from $b$ to $s\cdot b$. The scaling rule for the embedding layers is as follows. First, consider the case all ids are frequent, then doubling the batch size doubles the occurrence of these ids in the batch. Thus, the gradients are also doubled. This indicates a linear scaling on the gradient clipping value.

However, if all the ids are infrequent, considering the process of merging $s$ small batches into a big batch. For a specific id, the probability that it occurs in two of $s$ batches is small. Thus, with the assumption that no colliding ids occur in the $s$ small batches, we have the gradient $g'$ for the large batch $B'$ ($g$ for the small batch $B$):
\begin{equation*}
    \bm{g'} = \sqrt{\sum_{i=1}^s\bm{g}_s}=\sqrt{s}\bm{g}.
\end{equation*}
This indicates we should use a square root scaling for the gradient clipping value. In practice, although both frequent and infrequent ids exist, we find the sparse one dominates the gradients. We find that the scaling in the norm of gradients is approximately $\sqrt{s}$ when scaling the batch size and suggest square root scaling on gradient clipping value on the embedding layer is a better choice.

% First, consider the expected number of $\text{id}_k^{\text{f}_j}$ in batch $B$:
% \begin{equation*}
%     \E[|\text{id}_k^{\text{f}_j}\in B|]=\sum_{x\in B}\prob(\text{id}_k^{\text{f}_j}\in x)=b\cdot\texttt{freq}(\text{id}_k^{\text{f}_j})
% \end{equation*}
% Together with equation~(\ref{eq:case}), we know for the gradient of an frequent id embedding vector $\bm{g}$, the expected gradient is:
% \begin{equation*}
% \E[\bm{g}]=\E[|\text{id}_k^{\text{f}_j}\in B|]\nabla L(w,x)=b\cdot\texttt{freq}(\text{id}_k^{\text{f}_j})\cdot\nabla L(w,x)
% \end{equation*}
% As a result, we need to apply linear scaling.

% However, for the infrequent ids, $b\cdot\texttt{freq}(\text{id}_k^{\text{f}_j})$ is less than one even $b$ is enlarged, which means they do not appear in most batch. Thus, the number of gradients to that column is not increased when batch is scaled up. As shown in equation~(\ref{eq:case}), the possibility of infrequent ids showing up is scaled up, we expect that more unshown ids in a small batch now appear in a large batch. With the assumption that when merging $s$ small batches to a big batch, there is no colliding ids, we have the gradients:
% \begin{equation*}
%     \bm{g}
% \end{equation*}

\section{Proof Sketch for Convergence of CowClip}
\label{sec:proof}
To see the convergence property of our algorithm, we go through the following steps to get the CowClip optimizer, and the convergence of CowClip is ensured by each component.

First, the CowClip algorithm only applies to the embedding layers (the first layer) of the whole model. The convergence of the dense part is guaranteed by the Adam optimizer~\cite{Kingma2015AdamAM}. Next, if we change the gradient clipping operation in CowClip to gradient normalization, our algorithm can be viewed as a variant of the LAMB optimizer~\cite{You2020LargeBO}, which also ensures convergence. The differences between this version of CowClip and LAMB are as follows. CowClip focuses on a smaller granularity, which has been studied in the AGC optimizer~\cite{brock2021high} CowClip also scales the threshold by the occurrence time, but this is to correct the reduce-mean from each field's perspective. Finally, we relax CowClip from normalization to gradient clipping. The convergence of gradient clipping has been fully understood according to~\cite{Zhang2020WhyGC,mai2021stability}. A clipping method is a weak form of the normalization method (leaving the gradients below the threshold unchanged). Thus the relaxation does not change the convergence of the CowClip algorithm.

\section{Discussion and Future Work}
\label{sec:dis}
With CowClip, we can scale the batch size of the CTR prediction model to 128 times larger than the original size on one GPU. Despite the great power of our method, there are more works to be done for large batch CTR prediction training, which we leave as future works.

% multi-gpu (communication cost)
First, as mentioned in the introduction, many works have been devoted to designing a sound system for a multi-node CTR prediction model. Due to the computational resource, we verified our method on a single GPU setting. Although it seems straightforward to integrate our approach into a multi-GPU training setting, system optimization is still needed for fast distributed training. It is also interesting to know how much can our method accelerate the training in a communication and memory-efficient multi-node CTR prediction system.

% larger batch size
Second, when scaling to a very large batch size (\eg, 256K or even larger), the AUC still drops even with CowClip. One possible reason for this is that as the batch size grows, the assumption most ids are infrequent may not be held. One possible way to deal with the problem is to design an id-wise scaling strategy, which may not be computational-efficient. Another possibility lies in the loss of generalization ability of models trained at a large batch, as found and discussed in CV and NLP areas. Since our experiments are conducted on only one GPU, a larger batch size is needed when scaling to a~multi-GPU setting.

% adam and laze-l2
In addition, our experiments use the setting of adam optimizer and L2-regularization on all weights. In this setting, every weight and embeddings, along with their optimization states in the optimizer, are updated in each step. In some modern CTR prediction systems, a 'lazy' optimizer (\eg, adagrad~\cite{Duchi2010AdaptiveSM}, lazy-adam) updates the state of embeddings only when the corresponding id appears in the batch, and an L2-regularization only imposed on these ids is used for fast computation. In addition, sparse representation of embedding matrix also makes a difference to the optimization process (\eg, sparse optimizer update of tensor in Tensorflow). The clipping strategy should be modified to suit these methods. Combining our method with these variants of optimization strategies is also an interesting problem for practical deployment.

% one-epoch training
Our experiments follow previous work and train the network with 10 epochs. In reality, when the training dataset is huge, it is unaffordable or too slow to do multi-epoch training. In this case, one-epoch training is the choice to train an update-to-date CTR prediction model. As shown in the training curve in Figure~\ref{fig:test}, although we achieve better results when finishing the training, the AUC at the first epoch drops compared to the small-batch setting. We also find that to maintain the first epoch AUC value, a much smaller L2-regularization should be adopted. The reason may be that in the first training epoch, overfitting is not likely to occur and thus requires a weaker regularization. We believe an investigation into one-epoch large batch training will be valuable work.

% extend to other place need embedding
Apart from CTR prediction, there are other tasks with a large embedding table, such as NLP tasks. For instance, in the Chinese embedding table, an unbalanced-frequency exists among different characters. Even if the frequencies of different ids are not as varied as those in CTR prediction, a simplified version of CowClip (\eg, remove the occurrence count) may help to stabilize the training of models in these tasks.

With the growth of hardware and modern CTR prediction systems, we believe a trend is to adopt a larger and larger batch size for fast CTR prediction model training. To use a large batch size, apart from a robust system, a suitable algorithm is also needed to maintain the performance. As there are many works on large batch training in CV and NLP, few works discuss the problem of large batch training in the CTR prediction model. We think it is worthwhile to investigate this problem.

\section*{Broader Impact}
\label{sec:impact}
Accelerating the training speed of the CTR prediction model by large batch training is directly beneficial to the ad-tech and e-commerce practicians. Time and cost are reduced for re-training a model, contributing to faster product development iterations. In addition, the personalized recommendation could be more accurate and up-to-date, which potentially improves the user experience.

\end{document}